\pdfoutput=1

\documentclass[11pt]{article}

\usepackage[final]{acl}

\usepackage{times}
\usepackage{latexsym}

\usepackage[T1]{fontenc}

\usepackage[utf8]{inputenc}

\usepackage{microtype}
\usepackage{CJKutf8}
\usepackage[T1]{fontenc}
\usepackage{inconsolata}
\usepackage{longtable}
\usepackage{array}
\usepackage{tabularx}
\usepackage{graphicx}
\usepackage{booktabs}
\usepackage{arydshln} 
\usepackage{multirow}
\usepackage{kotex}
\usepackage{setspace}
\usepackage{multicol}
\usepackage{supertabular}
\usepackage{enumitem}
\usepackage{xcolor}
\usepackage{amsmath} 
\setlist[itemize]{nolistsep}
\title{BIPED: Pedagogically Informed Tutoring System for ESL Education}

\author{%
\textbf{Soonwoo Kwon}\textsuperscript{1}\footnotemark[2] \hspace{4mm}
\textbf{Sojung Kim}\textsuperscript{2}\footnotemark[2] \hspace{4mm}
\textbf{Minju Park}\textsuperscript{3} \hspace{4mm}
\textbf{Seunghyun Lee}\textsuperscript{3} \hspace{4mm}
\textbf{Kyuseok Kim}\textsuperscript{3}\ \\
{Twelve Labs}\textsuperscript{1} \hspace{4mm}{KAIST}\textsuperscript{2} \hspace{4mm}
{Riiid AI Research}\textsuperscript{3} \hspace{4mm} 
\\
\texttt{ian@twelvelabs.io\textsuperscript{1}, sojung\_kim@kaist.ac.kr\textsuperscript{2},} \\
\texttt{\{minju.park,seunghyun.lee\}@riiid.co\textsuperscript{3}, kimkyu80@gmail.com\textsuperscript{3}}
}

\begin{document}
\maketitle
\renewcommand{\thefootnote}{\fnsymbol{footnote}}
\footnotetext[2]{Work done while at Riiid}

\begin{abstract}
Large Language Models (LLMs) have a great potential to serve as readily available and cost-efficient Conversational Intelligent Tutoring Systems (CITS) for teaching L2 learners of English. 
Existing CITS, however, are designed to teach only simple concepts or lack the pedagogical depth necessary to address diverse learning strategies.
To develop a more pedagogically informed CITS capable of teaching complex concepts, we construct a BIlingual PEDagogically-informed Tutoring Dataset (\textbf{BIPED}) of one-on-one, human-to-human English tutoring interactions. 
Through post-hoc analysis of the tutoring interactions, we come up with a lexicon of dialogue acts (34 tutor acts and 9 student acts), which we use to further annotate the collected dataset. 
Based on a two-step framework of first predicting the appropriate tutor act then generating the corresponding response, we implemented two CITS models using GPT-4 and SOLAR-KO, respectively. 
We experimentally demonstrate that the implemented models not only replicate the style of human teachers but also employ diverse and contextually appropriate pedagogical strategies. 
\end{abstract}

\section{Introduction}
\label{sec:intro}

As Large Language Models (LLMs) such as GPT~\citep{achiam2023gpt} revolutionize the field of natural language generation, both researchers and practitioners have put an increasing amount of effort into developing Conversational Intelligent Tutoring Systems (CITS) that leverage the generative capabilities of LLM’s~\citep{tack2022ai, abdelghani2022gpt, minju, lee-etal-2023-peep}.
Specifically, LLMs have the potential to teach English as a Second/Foreign Language (ESL/EFL), for they may serve as readily-available tutors that can emulate native-speaking contexts~\citep{minju, lee-etal-2023-peep}.

However, most CITS proposed in literature for teaching ESL are restricted in scope and lack pedagogical depth.
For example, works based on CIMA dataset~\citep{stasaski2020cima} consider toy concepts of colored shapes and prepositions. 
Possibly due to limited scope, CIMA only considers a five tutor act classes (“hint”, “question”, “correction”, “confirmation”, and “other”). 
Similarly, the TSCC dataset~\citep{caines2022teacher} only considers abstract tutor act classes such as “scaffolding”, and does not consider a more granular set of teaching strategies (e.g., inferential clues, teaching synonyms or antonyms, etc.). 
Consequently, models trained on CIMA~\citep{stasaski2020cima} and TSCC~\citep{caines2022teacher} are limited to employing simplistic instructional strategies~\citep{macina2023opportunities}.


The goal of our work is to develop CITS that can teach ESL by applying pedagogically meaningful teaching strategies adaptively to student’s needs. 
To do so, we designed a tutoring session that focuses on comprehending a real news article, covering vocabulary, grammar, and relevant cultural contexts. 
Based on this, we construct a BIlingual PEDagogically-informed Tutoring Dataset (BIPED). 
For BIPED, we collected a bilingual, one-on-one human-to-human tutoring dataset.
Thereafter, we analyzed the dataset post-hoc from a pedagogical viewpoint and developed a categorization of dialogue acts, which comprises 34 tutor acts and 9 student acts. 
Finally, we annotated the data using the defined dialogue act categories.



As for the development of CITS, we employ the framework~\citep{macina2023opportunities, wang2023strategize} whereby the LLM first chooses the suitable tutor act, then generates the corresponding utterance. 
We believe this approach enables the model to generate a more focused response that does not deviate from the chosen tutor intent. 
We consider two implementations of such CITS, one based on GPT-4 prompting, and another based on fine-tuning SOLAR-KO 10.7B~\citep{solar_ko_junbum_2023}, which is a public model trained on a Korean corpus. For GPT-4, our dataset is used as examples for in-context learning. 
For fine-tuning, we explored a multi-task learning approach, where the model is instruction-tuned to carry out tutor act prediction and response generation, as well as other tasks for context grounding. 


In our experiments, we show through various metrics that the implemented models are capable of mimicking human teacher’s utterance style, as well as their pedagogical strategies. 
Specifically, we show that our models choose appropriate and diverse tutor acts, and generate human-like utterances, as measured by sentence similarity metrics. 


The contributions of our paper are as follows:
\begin{itemize}[leftmargin=*]
\item We provide BIPED, a dataset curated from bilingual human-to-human tutoring dialogues that come annotated with specific tutoring acts. Importantly, the lesson topic is complex enough to require sophisticated pedagogical methods.
\item We implement two CITS based on BIPED. In particular, we explore both in-context learning and fine-tuning approaches. It is the pioneering work in utilizing instruction tuning techniques in the development of CITS.
\item We provide experimental results demonstrating the implemented models' ability to mimic human tutors. 
\end{itemize}

\section{Related Work}
\label{sec:relatedwork}

\subsection{Conversational Intelligent Tutoring Systems (CITS)}

Some of conversational tutoring systems make effective use of GPT through the implementation of prompt engineering~\citep{minju, tack2022ai, abdelghani2022gpt}. Moreover, there exist studies that tailor open-source LLMs to fulfill distinct objectives~\citep{qian2023user,huang2022direct}, and the generation of dialogue in conjunction with act classification is also gaining ground~\citep{wang2023strategize}.
However, despite these advancements in CITS using LLMs, a perceivable lack in the pedagogical approach is observed~\citep{macina2023opportunities, kasneci2023chatgpt}.
These systems fall short of human tutors in emulating tutor's speech, comprehending students, and assisting them in an appropriate way~\citep{tack2022ai}.
In order to ensure that tutoring systems can effectively mimic the qualities of an actual tutor, the need for the integration of pedagogical attributes into the design of these systems is underscored~\citep{du2016modelling}.

\subsection{Tutoring Dialogue Dataset}
\begin{table*}[h]
\small{
\centering
\resizebox{\textwidth}{!}{
\begin{tabular}{cccccc}
\toprule

\textbf{Metric} 
& \textbf{TSCC v2} & \textbf{TalkMoves} & \textbf{NCTE} & \textbf{BIPED} \\ 
&\cite{caines2022teacher} &\cite{suresh2022talkmoves} & \cite{demszky2022ncte} & \\ 
\midrule
\# tutoring sessions & 260 & 567 & 1,660 & 112 \\
\# tutors & 2 & - & 317 & 4 \\
\# students & 13 & - & 10,817 & 60 \\
Avg. turns per session & 159.5 & 211.2 & 350 & 105.7 \\
Avg. words per turn	& T 10.8 / S 6.4 & T 8.2 / S 5.9	& T 29.1 / S 4.0 & T 31.7 / S 8.2 \\
\# dialogue acts & T,S 12 & T 6 / S 4 & T 3 / S 2 & \textbf{T 34 / S 9} \\
domain & ESL & math & math & ESL \\
setting & 1:1 & classroom & classroom & 1:1 \\ \bottomrule
\end{tabular}
\normalsize}
\caption{Comparison of BIPED with existing dialogue tutoring datasets: BIPED contains acts that are not only more diverse but also have greater pedagogical depth compared to previous datasets.}
\vspace{-0.3cm}
\label{tab:datasets}}
\end{table*}
In the field of general natural language processing, the procedure of identifying the dialogue act serves as the preliminary phase in modeling intricate discourse structures~\citep{stolcke2000dialogue,budzianowski2018multiwoz,saha2020towards}. 
Similarly, within the realm of education, related works also focus on dialogue acts within the tutoring dialogues~\citep{caines2022teacher, tan2023does, lin2023robust, nyehierarchical}.
In contrast to other areas, tutoring dialogue acts possess distinct characteristics in that they include the instructional intentions of tutors, with the objective of elevating the effectiveness of the tutoring process. There exist extensive studies dedicated to the classification of teaching strategies within the educational field~\citep{brown2015teaching,brown2014principles,walsh2006investigating,walsh2013classroom}.
While numerous studies of CITS draw from this established body of knowledge, they often encounter constraints.
Due to the complexity of incorporating an expansive range of pedagogical methods within CITS, numerous studies often narrow the scope to a limited set of teaching strategies~\citep{alic2022computationally, demszky2022ncte}, or alternatively, focus solely on teaching simple knowledge concepts which require a more confined set of instructional strategies~\citep{stasaski2020cima}. 
To address this limitation, our work endeavors to integrate a diverse array of teaching methodologies, devised in collaboration with domain experts.

There is a considerable variation across existing tutoring dialogue datasets in terms of its specific composition.
There are tutoring dialogues assembled from interactions between learners and automated systems~\citep{huang2022direct, macina2023mathdial}.
Meanwhile, \citealp{stasaski2020cima} have generated dialogues wherein experts simulate both the tutor and student roles.
Nonetheless, to authentically encapsulate the dynamic interactions between the tutor and the student, it becomes imperative to establish a real teaching context.
Moreover, focusing on ESL, there exists a significant deficit in the availability of bilingual data.
In the context of language learning, pedagogical strategies that simultaneously consider both languages – for instance, code-switching – play a pivotal role~\citep{brown2014principles}.
Thus, the shortage of bilingual data is restricting the efficacy and scope with respect to language instruction~\citep{nguyen2022building}.


Highlighting that our dataset, BIPED, is primarily focused on human-to-human tutoring dialogues including annotations, we provide an overview of descriptive statistics with a comparison to existing datasets in comparable settings in Table~\ref{tab:datasets}.

\begin{figure*}[h]
\centering
\includegraphics[width=\linewidth]{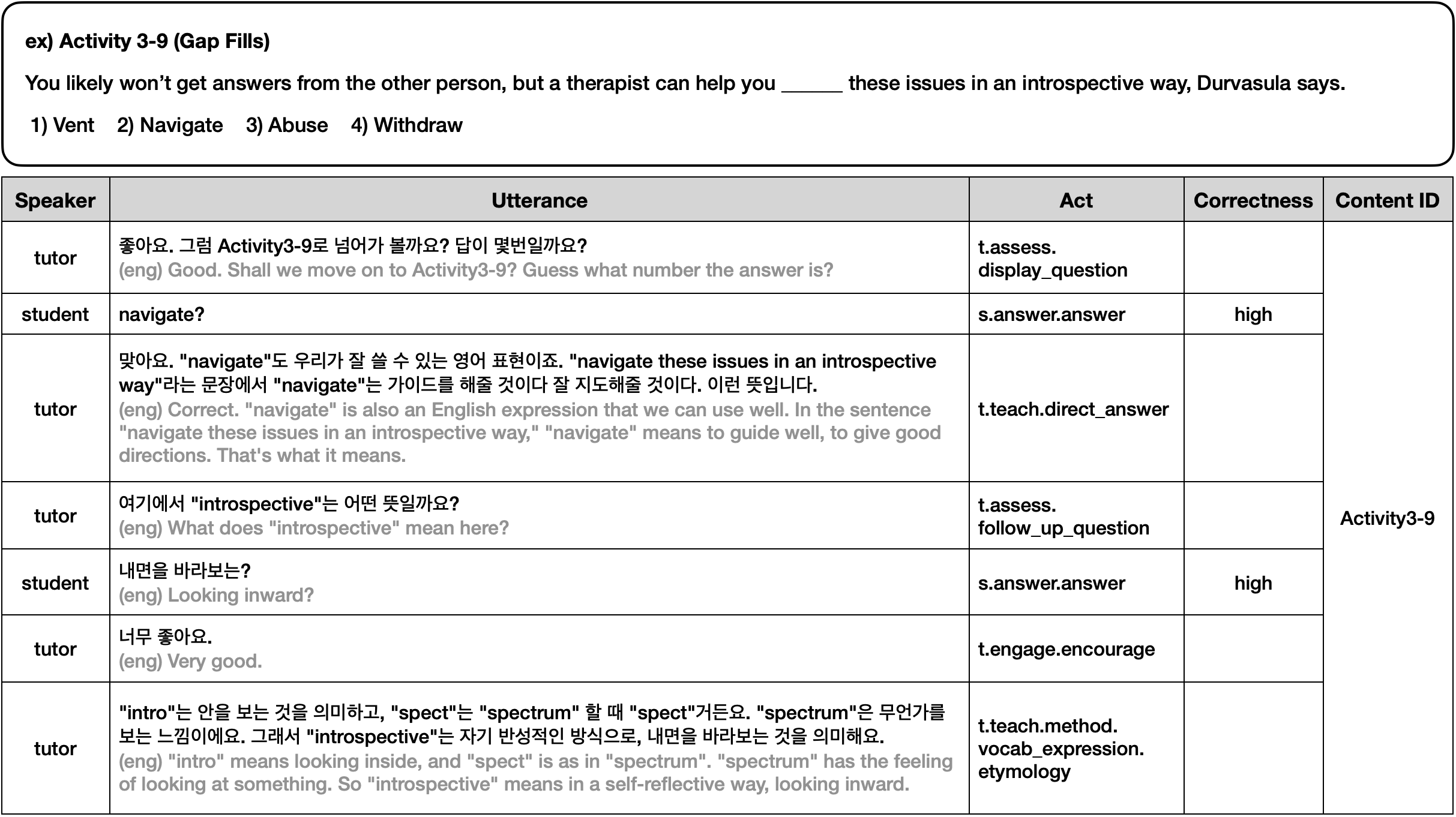}
\caption{Example of our dataset, BIPED. It includes a series of dialogues between a tutor and a student, annotated with dialogue acts, content information, and the correctness of student responses.}
\label{fig:data_sample}
\vspace{-1mm}
\end{figure*}

\section{Dataset: BIPED}\label{sec:dataset}

\begin{figure}[t]
\centering
\includegraphics[width=\columnwidth]{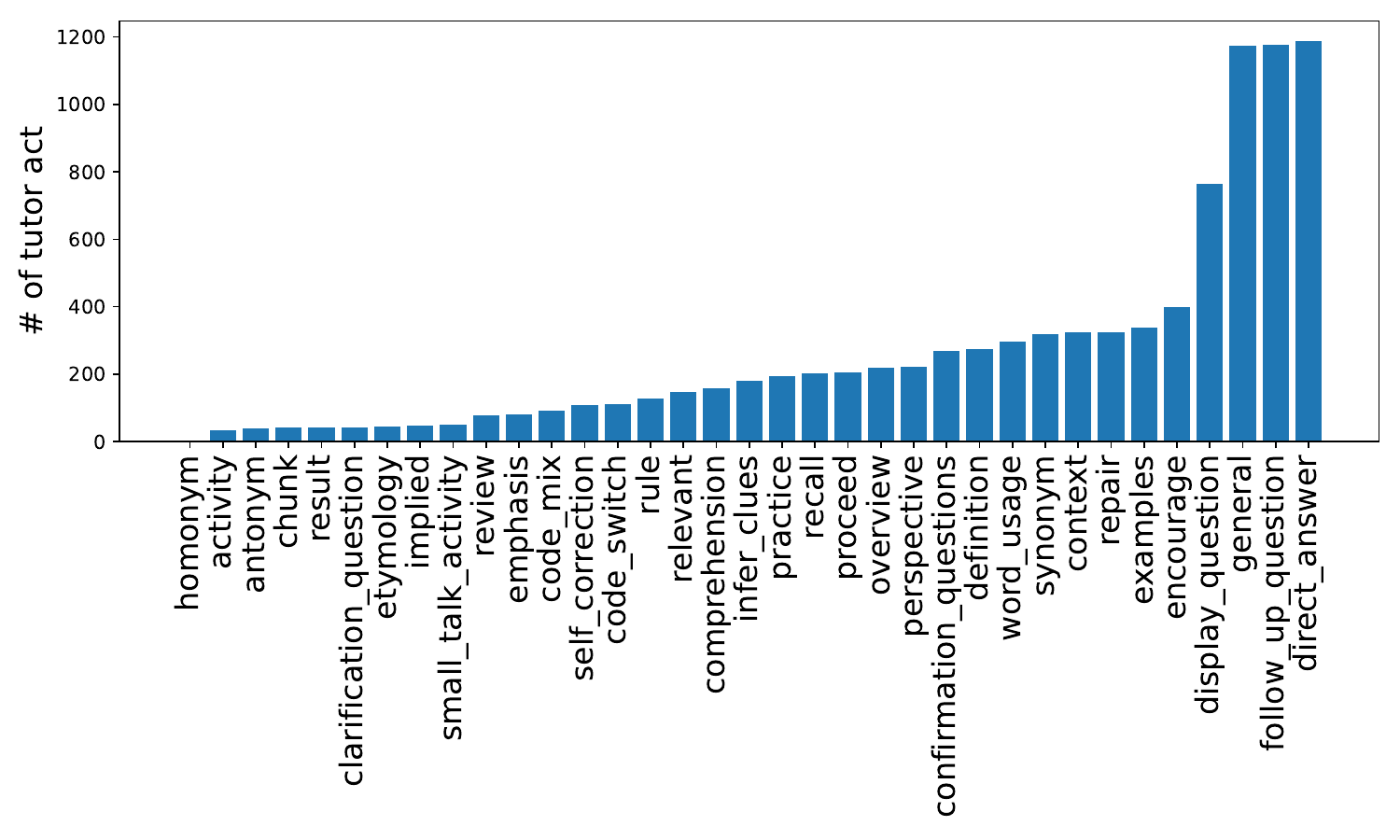}
\caption{Distribution of tutor acts in BIPED}
\vspace{-1mm}
\label{fig_data_dist}
\end{figure}

This section describes our dataset, BIPED, detailing its design, collection, annotation process, and essential statistics.

\subsection{Data Design}
\label{subsec:data_design}
We aim to construct a comprehensive tutoring dataset that contains the entire dialogues from tutoring sessions, with the corresponding dialogue acts and encompassing contents labeled on each utterance.
Additionally, we categorized the correctness of student responses to inquiries posed by the tutor.
To provide a clearer understanding of the structure of our dataset, a sample of the dataset is displayed in Figure~\ref{fig:data_sample}.

\subsection{Data Collection}
\label{subsec:data_collection}
For the data collection through one-to-one tutoring sessions, we recruited four experienced ESL tutors and 60 students with a willingness to pursue English studies. Every student was partnered with a single tutor, and over the course of three days, they had three successive 30-minute one-to-one tutoring sessions.
Prior to the first session, tutors were given an additional 10-minute to comprehend the student's individual status, including their particular objectives for studying English, their methods of studying English, and also their personal interests.
The learning contents, developed by subject matter experts in ESL teaching, consisted of an English article along with activities focused on vocabulary and reading comprehension. 
Based on the student's pretest results and the ongoing assessment of their progress throughout the class, including the preceded 10-minute talk, the tutors were required to adapt their teaching style to each unique learner, selectively determining the learning materials to cover.
The tutoring sessions were held in an audio-only setting, and the transcriptions of these sessions were obtained with the assistance of Whisper~\citep{radford2023robust}.

The dataset contains 112 tutoring sessions\footnote{We confine the utilization of data to a subset that has been fully annotated, despite a larger volume of collected data.}, as outlined in Table~\ref{tab:datasets}. On average, each session consisted of roughly 105.7 turns. Analysis reveals that the utterances made by tutors in a single turn contained an average of 31.7 words, while the student’s utterances were more concise, averaging 8.2 words per utterance.




In order to confirm the educational significance of the tutoring sessions, students were required to take a pre-test and post-test surrounding these sessions. 
Drawing upon the computation of the normalized learning gain, the average result was fount to be 0.089, demonstrating the tutoring sessions' effectiveness in enhancing the students' English proficiency.
A detailed explanation about the pre/post-test design and the evaluation of learning gain from these tests is described in Appendix~\ref{test_design} and Appendix~\ref{learning_gain}, respectively.
Additionally, a feedback survey was deployed to understand their perspectives and feelings towards the courses.
Overall, the feedback suggested a general contentment with the educational experience provided.
For instance, for the questions “To what extent did you find the class satisfactory?”, “How suitable for you was the teaching method employed by the tutor during the class?”, “Was the class conducted with consideration for your linguistic background and cultural understanding?”, the respondents indicated high approval with average scores of 4.65, 4.60, and 4.68 out of a possible 5, respectively. These results support the pedagogical excellence of our dataset.

\subsection{Dialogue Act Design}
\label{sec:dialog_act_design}
To accurately encapsulate the diverse intentions of both tutors and students within our dataset, we defined 34 tutor acts and 9 student acts in collaboration with experienced ESL instructors.
Our taxonomy was grounded in prior research~\citep{caines2022teacher, stasaski2020cima}, studies related to teaching methods~\citep{brown2014principles, brown2015teaching}, and underwent further refinements after observing multiple pilot classes.
As the tutor acts play a crucial role in influencing the efficacy of a tutoring session, we have accorded significant importance to the process of categorizing tutor acts. Tutor acts have been classified into five primary categories:
\begin{itemize}
  \item \textit{General}: general utterances including greeting, farewell, small talk, short reply, etc.
  \item \textit{Operational}: utterances designed to facilitate the smooth operation of the class.
  \item \textit{Assessment}: utterances intended to evaluate the student's learning progress.
  \item \textit{Teaching}: utterances dedicated to teach a certain knowledge to the student.
  \item \textit{Engagement}: utterances tailored to encourage student participation and involvement.
\end{itemize}
The respective categories encompass 1,3,4,22, and 4 distinct tutor acts.
Prior research has primarily concentrated on broader behaviors, and current language models have shown proficiency in their application to such contexts. 
Given the limitations observed within current taxonomies in terms of capturing the diversity of teaching methods, our study endeavors to provide a more detailed examination of the "Teaching" category, while drawing upon preliminary studies in reference to the remaining categories. The detailed process for developing the tutor act taxonomy is described in Appendix~\ref{taxonomy}.

Concurrently, in a more simplified manner, student acts are divided into four distinct categories: \textit{General}, \textit{Operational}, \textit{Question}, and \textit{Answer}. These particular categories encompass 1, 3, 2, and 3 unique student acts respectively.
The complete list of both tutor and student acts,  along with corresponding examples is exhibited in the Appendix~\ref{tutor_act} and \ref{student_act}.

\subsection{Data Annotation}

\subsection*{Annotation Process}
Initially, we implemented a set of processing steps, including the anonymization and correction of grammatical errors.
Subsequently, with the taxonomy design described in the Section~\ref{sec:dialog_act_design}, each utterance were annotated with the relevant act labels by our trained human annotators.
In cases where a single utterance encompassed multiple acts, the utterance was split to ensure that each utterance corresponded to a single action. 
These segmented utterances are referred to as \textit{act-level utterances}. 
That is, a single turn might consist of several act-level utterances. Our observations revealed that, on average, there were 145 act-level utterances per session, with tutors contributing 83.2 and students contributing 53.1. This indicates that tutors employed a variety of actions more frequently within the session. The distribution of tutor acts is presented in Figure~\ref{fig_data_dist}. It is noteworthy that the dataset exhibits a wide range of diverse acts, enhancing the pedagogical richness of our dataset.

\subsection*{Annotation Quality and Analysis}
To validate the annotation quality, we investigated the annotation agreement among annotators. We sampled 495 utterances from 4 tutoring sessions, consisting of 313 utterances from teachers and 182 from students. Three independent were tasked with annotating the tutor acts and the student acts. The overall annotator agreement, measured using Fleiss’s Kappa, was 0.70. Notably, despite a significant number of tutor acts(34), the agreement result regarding tutor acts was substantial, with a score of 0.64. Similarly, the agreement regarding student acts was 0.72, indicating a high level of concordance. The most disagreed act pair among teaching methods, denoted as "t.teach.method" in Appendix~\ref{tutor_act}, was “t.teach.method.vocab\_expression.word\_usage” and “t.teach.method.vocab\_expression.context”. This disparity likely arises from their shared focus on the contextual usage of vocabulary expression within definitions. Also, it was noted that there are instances where acts, which could be annotated with more specific labels within the “t.teach.method”, are instead categorized under “t.teach.direct\_answer”.

\section{Framework}
\label{sec:framework}
\label{subsec:motivation}
To build CITS that strategically incorporates pedagogical considerations, we follow the framework~\citep{macina-etal-2023-opportunities, wang-etal-2023-strategize} where the LLM (1) chooses the appropriate tutor act, (2) then generates the response accordingly. 
We advocate this two-step approach for the following two reasons.

\subsection*{Interpretability}
First, the proposed two-step framework allows us to better understand the intention of the model (tutor acts) when generating a certain response for a given dialogue context. 
As a consequence, this enables us to more rigorously evaluate the model from a pedagogical perspective. 
Specifically, human experts may now assess whether the model correctly selected the appropriate type of act, and whether the generated response is suitable.

\subsection*{Controllability}
More importantly, the proposed framework allows for greater controllability of CITS. 
By conditioning the LLM to a specific tutor act class, we ensure more focused response generation without deviation from the target act.
We expect this to be essential when generating more sophisticated responses (e.g., teaching etymology, code-mixing), for these are sparsely distributed compared to more generic responses such as simple direct answer as shown in Figure~\ref{fig_data_dist}.

Moreover, a crucial aspect for controllability is the model's amenability to improvement via continual refinement. 
Our framework allows for a divide-and-conquer approach for model refinement; namely, given dialogue context and student response, instead of requiring the human expert to generate a full response, we may let them simply label the best teacher response class (for instruction tuning), or a ranked list of teacher response classes (for RLHF). 
As for improving response generation, we may consider sophisticated data augmentation schemes, where we synthesize responses for a given tutor act class based on retrieved samples of the same class. 
\begin{table*}[ht]
\renewcommand{\arraystretch}{1.1}
\centering

\small 
{\begin{tabular}{p{0.18\linewidth}p{0.75\linewidth}}
\hline
\textbf{Type} & \textbf{Instruction}\\ \hline
(1) Act prediction & 
\textbf{\#\#\# Instruction:} Select an appropriate next tutor act based on the given 'dialogue context' and 'act candidates'. In the context, tutor uses various tutor act and contents(activities or passages), depending on the student's abilities and questions. Tutor act represents a teaching method that you can exploit. Choose the best act catering to the context and understanding the meaning of each act."  \newline
\textbf{\#\#\# Input:}  \newline - Context: \{\textit{dialogue context (Appendix~\ref{context_example})}\}, \newline - Act candidates: \{\textit{act candidates (Appendix~\ref{tutor_act}})\} \newline 
\textbf{\#\#\# Response:} t.teach.method.vocab.expression.etymology \\
\hline
(2) Utterance generation & 
\textbf{\#\#\# Instruction:} As an English tutor for a Korean student, your job is to create coherent and tailored responses based on the given 'dialogue context' and 'tutor’s act'. In the context, tutor uses various teaching acts and contents(activities or passages), depending on the student's abilities and questions. The given action is a teaching method that you have to exploit. \newline 
\textbf{\#\#\# Input:} \newline  - Context: \{\textit{dialogue context (Appendix~\ref{context_example})}\}, \newline - Act: t.teach.method.vocab.expression.etymology, Explain a word or a phrase based on its etymology. When possible, discuss the prefix and suffix of the word. \newline
\textbf{\#\#\# Response:} "intro"는 안을 보는 것을 의미하고, "spect"는 "spectrum"할 때 "spect"거든요. "spectrum"은 무언가를 보는 느낌이에요. 그래서 "introspective"는 자기 반성적인 방식으로, 내면을 바라보는 것을 의미해요. \textcolor{gray}{(eng) "Intro" means looking inward. "spect" is same as "spect" in "Spectrum" and feels like seeing something. So, "introspective" means in a self-reflective manner, looking inward.} \\

\hline
(3) Missing context inference & \textbf{\#\#\# Instruction:} Predict a speaker and their utterance to fit between the given 'dialogue context' of a student and a teacher and the final 'utterance' made by the teacher. \newline \textbf{\#\#\# Input:} \newline - Context:   \{\textit{partial context (Appendix~\ref{context_example})}\}, \newline - Utterance: "intro"는 안을 보는 것을 의미하고, "spect"는 "spectrum"할 때 "spect"거든요. "spectrum"은 무언가를 보는 느낌이에요. 그래서 "introspective"는 자기 반성적인 방식으로, 내면을 바라보는 것을 의미해요. \textcolor{gray}{(eng) "Intro" means looking inward. "spect" is same as "spect" in "Spectrum" and feels like seeing something. So, "introspective" means in a self-reflective manner, looking inward.} \newline \textbf{\#\#\# Response:} tutor, 너무 좋아요. \textcolor{gray}{(eng) Very good.} \\ 
\hline
(4) Minority act utterance generation & \textbf{\#\#\# Instruction:} As an English teacher for Korean students, generate tutor's utterance to teach the given learning content with the act. \newline \textbf{\#\#\# Input:} \newline - Content: [Activity 3-9] You likely won’t get answers from the other person, but a therapist can help you \_\_\_ these issues in an introspective way, Durvasula says. (1) Vent (2) Navigate (3) Abuse (4) Withdraw \newline - Act: t.teach.method.vocab.expression.etymology, Explain a word or a phrase based on its etymology. When possible, discuss the prefix and suffix of the word. \newline \textbf{\#\#\# Response:}  "Intro"는 "안쪽의"라는 뜻이고, "spect"는 "보다"라는 뜻입니다. 그래서 "Introspective"는 "자아성찰하는"이라는 뜻입니다. \textcolor{gray}{(eng) "Intro" means "inner" and "spect" means "to see." So, "introspective" means "self-reflective."}\\ 
\hline
\end{tabular}}
\caption{Proposed instruction types for fine-tuning. Examples of input formats are illustrated in the appendix.}
\label{instruction_example}
\end{table*}

\section{Method}
\label{sec:method}
In this section, we introduce two concrete implementations of the framework proposed in Section~\ref{sec:framework}.  
First implementation leverages GPT-4's in-context learning ability and is easily implemented via prompting,
while the second implementation instruction-tunes open-sourced LLM to perform various relevant tasks.  
\subsection{GPT-4 Prompting}
We first consider a GPT-4-based implementation via extensive prompting.  
While this implementation does not require model training, it is costly and requires a large amount of tokens for in-context learning. 

As outlined in Appendix \ref{prompt}, our prompt structure adheres to our framework. The first step involves choosing an effective tutor act from a provided tutor acts list. Next, an utterance that aligns with this chosen tutor act is generated. These steps consider both dialogue context and learning content provided in the test set. For few-shot learning, an example scenario from the training set that matches the tutor act in the task scenario is also provided. This includes dialogue context, learning content, a chosen tutor act, and a generated utterance.

\subsection{Instruction Fine-tuning}
We consider instruction-tuning a pre-trained LLM on the dataset introduced in Section~\ref{sec:dataset}. 
Our goal is to demonstrate that, with properly curated datasets and appropriate fine-tuning methods, we may enable open-source models to rival GPT-4's performance.
We instruction-tuned open-sourced LLM on four sub-tasks, each of which we detail below.

\subsubsection*{Task 1. Act prediction}
To make our model select the most effective teaching strategy, we task the model to predict the next tutor act based on given learning content and dialogue context.
Given a dialogue in the training dataset, we created instruction samples for every dialogue turn of the tutor.

\subsubsection*{Task 2. Utterance generation}
As a conversational tutor for Korean L2 learners of English, the model must be able to generate responses that are (1) fluent in both Korean and English; and (2) contextually relevant and matching the tutor act the response is conditioned on.   
To achieve this, we devise a task of generating the tutor's response based on learning content, dialogue history, and the conditioning tutor act.
By training the model to imitate expert human tutor's responses, we expect the model to obtain bilingual fluency, as well as the ability to generate appropriate response based on the aforementioned conditioning information. 

\subsubsection*{Task 3. Missing context inference}
We found out that, trained on Task 1-2, the model often generated responses that are not properly grounded in the recent dialogue context.
This comes from the inherent task difficulty: compared to simple conversational tasks, teaching often requires a purposeful and focused long-context generation, where one small digression may hurt the overall quality of the generated response.
To address this issue, we devise the following task:
Given learning content, dialogue history, and corresponding teacher response, we remove this and that. 
The model is then tasked to predict the missing context given the rest of the information. 
This way, we expect the model's utterance to be locally more consistent.  

\subsubsection*{Task 4. Minority act utterance generation}
Finally, we notice that there are minority tutor act classes that are, despite their pedagogical important, sparsely distributed in the training dataset.
Therefore, to make sure the model understand these act classes better, we additionally collected the following data - for a given learning content and a selected group of minority act classes, we asked experts to provide plausible utterances for the corresponding act classes. 
Then, we tasked the model to generate these utterances conditioned on the corresponding learning content and act class. 
Although this additional dataset lack specific dialogue context, we expect it to guide the model in generating utterances for minority act classes. 



\section{Experiments}
\label{sec:experiments}
\begin{table*}[h]
    \centering
    \resizebox{0.9\textwidth}{!}{
    \begin{tabular}{c|cc|cccc}
    \toprule
     & \multicolumn{2}{c|}{Tutor Act}      & \multicolumn{4}{c}{Utterance} \\ 
    Metric & Accuracy$\uparrow$ & Invariability$\downarrow$   & sBLEU$\uparrow$  & BERTScore$\uparrow$ & Embedding sim.$\uparrow$ & Length  \\ \midrule
    Base GPT  & 0.155 &    6.955         & 1.802  & 0.655 & 0.635     & 80 $\pm$ 31 \\ \hdashline
    GPT-3.5 (0-shot) & - &    -         & 1.849  & 0.646 & 0.661    & 45 $\pm$ 36 \\ 
    GPT-3.5 (1-shot) & -  &    -      & 2.423  & 0.658   & \textbf{0.704}  & 45 $\pm$ 34 \\ 
    GPT-4 (0-shot) & -    &    -      & 1.687  & 0.659  & 0.668    & 66 $\pm$ 28 \\ 
    GPT-4 (1-shot) & 0.250  &   7.682           & 2.608  & 0.669    & 0.686  & 58 $\pm$ 29 \\ \hdashline
    Finetuned & \textbf{0.259} &  \textbf{5.273}          & \textbf{15.874}  & \textbf{0.716}   & 0.680  & 39 $\pm$ 52 \\ 
    Finetuned (ablation) & -     &  -     & 14.199  & 0.708    & 0.673  & 34 $\pm$ 20 \\ 
    Finetuned (baseline)  & -     &  -     & 14.426  & 	0.689    & 0.670  & 25 $\pm$ 25 \\
    \hdashline
    Ground-truth & 1.0   &   & 100.  & 1. & 0.707 & 27 $\pm$ 22\\ \bottomrule
    \end{tabular}}
    \caption{Results of prompted GPTs and finetuned models.}
    \label{main_table}
    \vspace{-3mm}
\end{table*}

\begin{figure*}[!]
\centering
\includegraphics[width=\linewidth]{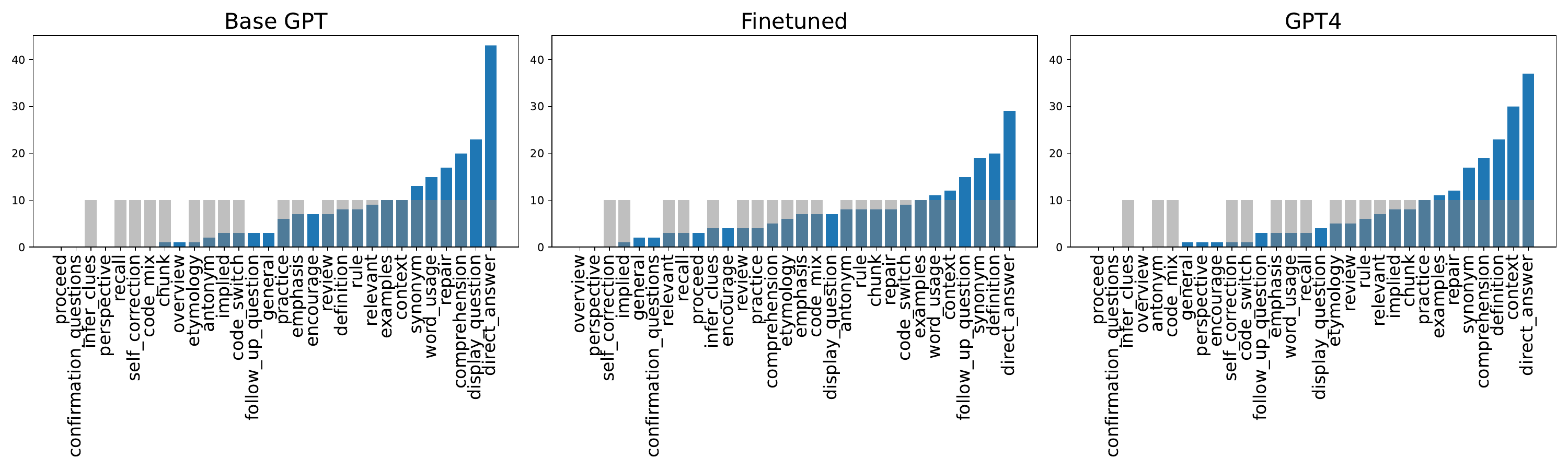}
\caption{The comparative distributions of chosen tutor acts. The grey bars represent the distribution of tutor acts in the test set while the blue bars denote the distribution of chosen tutor acts by Base GPT, our fine-tuned model, and GPT-4 (1-shot).}
\vspace{-3mm}
\label{fig:dist}
\end{figure*}

\subsection{Dataset split}

The experiment involved dividing BIPED randomly into training and test sets, organised by session. This was done based on the distribution of tutors within the dataset, allocating two sessions to one tutor, and one session each to the remaining tutors.  BIPED of 112 tutoring sessions was thus split into 102 training sessions and 10 test sessions. 

Extra data for generating minority act utterances, part of Task 4, were only included in the training set. The entire training data was reformulated into different instruction-tuning formats for each task.

The test set was devised into 220 scenarios, 10 unique scenarios for every one of the 22 tutor acts in the teaching category. Each test scenario was shaped around an act-level utterance initiated by a human tutor applying the respective tutor act, supported by the associated dialogue context and learning content. 

\subsection{Evaluation metric}
We conducted evaluations of the models based on our framework, focusing on two main aspects. Firstly, we evaluated the accuracy and invariability of tutor act predictions. For accuracy, we assessed the rate at which the model's predicted tutor act aligned with the actual act selected by the tutor in the test set. For invariability, we noted the frequency of each selected tutor act and calculated the mean absolute difference from a target count of 10. 

Secondly, we assessed the generated utterances. To measure the similarity between the generated and target tutor utterances, we utilized the sacrebleu implementation of BLEU(sBLEU)~\citep{post-2018-call} and BERT Score~\citep{Zhang*2020BERTScore:} using bert-base-multilingual-cased checkpoint. We note that the fine-tuned model sequentially predicted tutor acts and generated utterances based on these predictions. Additionally, we measured the coherence between generated utterances and previous ones in the given dialogue context using embedding similarity, specifically relying on OpenAI's text embedding(text-embedding-3-large). 

\subsection{Training details}
For fine-tuning we utilize SOLAR-KO-10.7B(~\citep{solar_ko_junbum_2023}), a model further trained on a Korean corpus to enhance bilingual comprehension and fluency. We adopt AdamW optimizer ~\citep{loshchilov2018decoupled} with a learning rate of 2e-5 and a batch size of 64. The maximum length is 4096. We train the model using 8 V100 GPUs with deepspeed Zero-3 offload for 2 epochs.

\subsection{Results}

The findings from our experiment, illustrated in the Table~\ref{main_table}, are compared against various models ranging from base GPT to fine-tuned models. 
Ground-truth represents the reformulated data gathered from human tutoring present in the test set.
Base GPT in Table~\ref{main_table} corresponds to the GPT-4 model bereft of any tutor act classification or our human tutoring examples, where its prompt is given in Appendix \ref{prompt}. We have also drawn a comparison involving GPT-3.5(gpt-3.5-turbo-16) and GPT-4 in a zero-shot and one-shot configuration. While they all adhere to the identical prompt structure, (as seen in Appendix \ref{prompt}), they diverge in aspects such as model compatibility and whether an example scenario is included. 
Lastly, we compared our two fine-tuned models. Fine-tuned refers to our instruction-tuned model, discussed in the method section~\ref{sec:method}, which was trained with instruction tasks 1 through 4. 
Conversely, 'Fine-tuned (ablation)' refers to the model trained with instruction tasks 1 and 2 only. 
We also included a comparison with a potential baseline, denoted as 'Fine-tuned (baseline)'. 
This model was trained to sequentially generate the following tutor act and corresponding utterance, conditioned on the given dialogue context without any instructions. This approach is similar to MATHDIAL~\citep{macina2023mathdial}, where the tutoring model was trained to generate utterances given the context and tutor act only. 

In assessing the tutor act criteria, our fine-tuned model outperforms other models in both accuracy and invariability as shown in Table \ref{main_table}. Although the accuracy of 0.259 may initially appear low, it's important to consider the complexity of choosing from amongst 34 distinct tutor acts. Moreover, a lower invariability score of 5.273 is indicative of our fine-tuned model's ability to utilize a wider array of tutor acts, thereby demonstrating more diverse pedagogical approaches. While the invariability score provides one aspect of variability, Figure \ref{fig:dist} presents a distribution of chosen tutor acts for the comparative models, which offers a more comprehensive view. Clearly, our fine-tuned model exhibits the greatest number of distinct tutor act classes. 

With respect to utterance-related metrics, our fine-tuned model delivers outstanding performances. It records the highest sBLEU score at 15.874 and BERTScore at 0.716, demonstrating its commendable ability in generating utterances that closely mirror the target tutor utterances. While the embedding similarity score of our fine-tuned model is slightly lower than that of GPT-3.5 (1-shot), it still exhibits higher coherence between individual utterances and their corresponding dialogue context. 

In terms of the length of the generated utterances, our model closely mimics the length of tutor utterances, whereas the general GPT models often produce unnecessarily long responses. We interpret this as positive, based on the assumption that the appropriate length of a tutor's utterance plays a vital role in maintaining student engagement.

Lastly, the fine-tuned model, trained with tasks 1 through 4, has outperformed the ablation model, which was trained with only partial instruction tasks, and the baseline model, which used simple conditional generation, across all evaluated metrics. This outcome highlights the added value of integrating additional instruction tasks (i.e., tasks 3 and 4) for improved tutoring performance. 

\subsection{Example of Interactive Evaluation}\label{sec:ie}

Our results primarily focus on static evaluation, where we predict the next tutor act and generate a corresponding response in one step for the test scenarios. However, a complete evaluation of the tutoring system, developed using our proposed models, necessitates interactive assessment with human students. This aspect of evaluation will be subjected to further study. Regardless, we offer a sneak peak into how an interaction plays out between a human and two tutoring models, namely GPT-4 (1-shot) and our fine-tuned model. 

The dialogue examples are included in Appendix \ref{interactive_eval}. 
Generally, when interacting with our tutoring systems, GPT-4 (1-shot) tends to produce lengthier responses, focusing mainly on transferring knowledge to students. Conversely, our fine-tuned model generates concise responses and actively promotes student participation. An interesting observation is that when a student insists that an incorrect answer is correct, GPT-4 maintains its answer, yet appears to accept the student's assertion (t.teach.emphasis). In contrast, our fine-tuned model re-emphasizes the correct answer (t.teach.repair), thereby highlighting the student's error.

\section{Conclusion}
\label{sec:conclusion}

In this study, we demonstrated the effectiveness of our proposed models in developing conversational intelligent tutoring systems for ESL learners. They excelled in replicating the utterance style of human tutors and implementing various and appropriate pedagogical strategies. Much of this improvement can be attributed to our newly constructed Bilingual and Pedagogically-Informed Tutoring Dataset (BIPED). Additionally, our two-step framework and the use of four instruction tuning tasks further contributed to the models' improvement.



\section{Limitations}
\label{sec:limitations}

Despite the encouraging results, our study has some limitations. First, our project lacks an interactive evaluation, left as future research direction. A formal evaluation of complete tutoring systems through human study would provide useful insights into their efficacy. 
Second, we were unable to utilize larger parameter models like LLaMa 70b \cite{touvron2023llama}, which could potentially improve the teaching capabilities of CITS. This restriction is worth considering for future studies, particularly with advancing computational resources and technologies.
Finally, the use of diverse instruction tuning tasks for multi-task modeling, using existing datasets and data augmentation, is not covered in this study. We believe this aspect carries opportunities for more nuanced model development and will be explored in subsequent work.

\section{Potential Risks}

Our research holds great promise, but it's also important to acknowledge potential risks. Firstly, Large Language Models might inadvertently teach misinformation due to their susceptibility to the 'hallucination' phenomenon, which generates information that isn't grounded in real-world facts. Secondly, there's risk of over-reliance on these intelligent tutoring systems. While they can play a pivotal role in language learning, they can't entirely mimic human tutors, especially in terms of adaptive strategies and emotional intelligence. 

\section*{Acknowledgment}
We thank our four dedicated ESL tutors, 60 students, and pilot study participants for their invaluable contributions. Special thanks to Yena Zhang for expertly managing all tutoring sessions and the data collection procedure. We also appreciate the annotators for their detailed work on our dataset, and the reviewers for their constructive comments, which greatly improved the quality of this paper.

\bibliography{main}

\begin{thebibliography}{37}
\expandafter\ifx\csname natexlab\endcsname\relax\def\natexlab#1{#1}\fi

\bibitem[{Abdelghani et~al.(2022)Abdelghani, Wang, Yuan, Wang, Sauz{\'e}on, and Oudeyer}]{abdelghani2022gpt}
R~Abdelghani, YH~Wang, X~Yuan, T~Wang, H~Sauz{\'e}on, and PY~Oudeyer. 2022.
\newblock Gpt-3-driven pedagogical agents for training children’s curious question-asking skills.
\newblock \emph{arXiv}, 2211.

\bibitem[{Achiam et~al.(2023)Achiam, Adler, Agarwal, Ahmad, Akkaya, Aleman, Almeida, Altenschmidt, Altman, Anadkat et~al.}]{achiam2023gpt}
Josh Achiam, Steven Adler, Sandhini Agarwal, Lama Ahmad, Ilge Akkaya, Florencia~Leoni Aleman, Diogo Almeida, Janko Altenschmidt, Sam Altman, Shyamal Anadkat, et~al. 2023.
\newblock Gpt-4 technical report.
\newblock \emph{arXiv:2303.08774}.

\bibitem[{Alic et~al.(2022)Alic, Demszky, Mancenido, Liu, Hill, and Jurafsky}]{alic2022computationally}
Sterling Alic, Dorottya Demszky, Zid Mancenido, Jing Liu, Heather Hill, and Dan Jurafsky. 2022.
\newblock Computationally identifying funneling and focusing questions in classroom discourse.
\newblock \emph{arXiv:2208.04715}.

\bibitem[{Brown(2014)}]{brown2014principles}
H~Douglas Brown. 2014.
\newblock \emph{Principles of language learning and teaching: A course in second language acquisition}.
\newblock Pearson.

\bibitem[{Brown and Lee(2015)}]{brown2015teaching}
H~Douglas Brown and Heekyeong Lee. 2015.
\newblock \emph{Teaching by Principles: An Interactive Approach to Language Pedagogy}.
\newblock P. Ed Australia.

\bibitem[{Budzianowski et~al.(2018)Budzianowski, Wen, Tseng, Casanueva, Ultes, Ramadan, and Ga{\v{s}}i{\'c}}]{budzianowski2018multiwoz}
Pawe{\l} Budzianowski, Tsung-Hsien Wen, Bo-Hsiang Tseng, Inigo Casanueva, Stefan Ultes, Osman Ramadan, and Milica Ga{\v{s}}i{\'c}. 2018.
\newblock Multiwoz--a large-scale multi-domain wizard-of-oz dataset for task-oriented dialogue modelling.
\newblock \emph{arXiv:1810.00278}.

\bibitem[{Caines et~al.(2022)Caines, Yannakoudakis, Allen, P{\'e}rez-Paredes, Byrne, and Buttery}]{caines2022teacher}
Andrew Caines, Helen Yannakoudakis, Helen Allen, Pascual P{\'e}rez-Paredes, Bill Byrne, and Paula Buttery. 2022.
\newblock The teacher-student chatroom corpus version 2: more lessons, new annotation, automatic detection of sequence shifts.
\newblock In \emph{Swedish Language Technology Conference and NLP4CALL}, pages 23--35.

\bibitem[{Demszky and Hill(2022)}]{demszky2022ncte}
Dorottya Demszky and Heather Hill. 2022.
\newblock The ncte transcripts: A dataset of elementary math classroom transcripts.
\newblock \emph{arXiv:2211.11772}.

\bibitem[{Du~Boulay and Luckin(2016)}]{du2016modelling}
Benedict Du~Boulay and Rosemary Luckin. 2016.
\newblock Modelling human teaching tactics and strategies for tutoring systems: 14 years on.
\newblock \emph{International Journal of Artificial Intelligence in Education}, 26:393--404.

\bibitem[{Hake(1998)}]{hake98}
Richard Hake. 1998.
\newblock \href {https://doi.org/10.1119/1.18809} {Interactive-engagement versus traditional methods: A six-thousand-student survey of mechanics test data for introductory physics courses}.
\newblock \emph{American Journal of Physics - AMER J PHYS}, 66.

\bibitem[{Huang et~al.(2022)Huang, Lee, and Kwon}]{huang2022direct}
Jin-Xia Huang, Yohan Lee, and Oh-Woog Kwon. 2022.
\newblock Direct: Toward dialogue-based reading comprehension tutoring.
\newblock \emph{IEEE Access}, 11:8978--8987.

\bibitem[{Kasneci et~al.(2023)Kasneci, Se{\ss}ler, K{\"u}chemann, Bannert, Dementieva, Fischer, Gasser, Groh, G{\"u}nnemann, H{\"u}llermeier et~al.}]{kasneci2023chatgpt}
Enkelejda Kasneci, Kathrin Se{\ss}ler, Stefan K{\"u}chemann, Maria Bannert, Daryna Dementieva, Frank Fischer, Urs Gasser, Georg Groh, Stephan G{\"u}nnemann, Eyke H{\"u}llermeier, et~al. 2023.
\newblock Chatgpt for good? on opportunities and challenges of large language models for education.
\newblock \emph{Learning and individual differences}, 103:102274.

\bibitem[{{L. Junbum}(2024)}]{solar_ko_junbum_2023}
{L. Junbum}. 2024.
\newblock \href {https://huggingface.co/beomi/SOLAR-KO-10.7B} {Solar-ko-10.7b}.

\bibitem[{Lee et~al.(2023)Lee, Jang, Park, Lee, Seo, Moon, Eo, Lee, Yahya, and Lim}]{lee-etal-2023-peep}
Seungjun Lee, Yoonna Jang, Chanjun Park, Jungseob Lee, Jaehyung Seo, Hyeonseok Moon, Sugyeong Eo, Seounghoon Lee, Bernardo Yahya, and Heuiseok Lim. 2023.
\newblock \href {https://doi.org/10.18653/v1/2023.acl-demo.18} {{PEEP}-talk: A situational dialogue-based chatbot for {E}nglish education}.
\newblock In \emph{Proceedings of the 61st Annual Meeting of the Association for Computational Linguistics (Volume 3: System Demonstrations)}, pages 190--207, Toronto, Canada. Association for Computational Linguistics.

\bibitem[{Lin et~al.(2023)Lin, Tan, Nguyen, Lang, Du, Buntine, Beare, Chen, and Ga{\v{s}}evi{\'c}}]{lin2023robust}
Jionghao Lin, Wei Tan, Ngoc~Dang Nguyen, David Lang, Lan Du, Wray Buntine, Richard Beare, Guanliang Chen, and Dragan Ga{\v{s}}evi{\'c}. 2023.
\newblock Robust educational dialogue act classifiers with low-resource and imbalanced datasets.
\newblock In \emph{International Conference on Artificial Intelligence in Education}, pages 114--125. Springer.

\bibitem[{Loshchilov and Hutter(2019)}]{loshchilov2018decoupled}
Ilya Loshchilov and Frank Hutter. 2019.
\newblock \href {https://openreview.net/forum?id=Bkg6RiCqY7} {Decoupled weight decay regularization}.
\newblock In \emph{International Conference on Learning Representations}.

\bibitem[{Macina et~al.(2023{\natexlab{a}})Macina, Daheim, Chowdhury, Sinha, Kapur, Gurevych, and Sachan}]{macina2023mathdial}
Jakub Macina, Nico Daheim, Sankalan~Pal Chowdhury, Tanmay Sinha, Manu Kapur, Iryna Gurevych, and Mrinmaya Sachan. 2023{\natexlab{a}}.
\newblock Mathdial: A dialogue tutoring dataset with rich pedagogical properties grounded in math reasoning problems.
\newblock \emph{arXiv:2305.14536}.

\bibitem[{Macina et~al.(2023{\natexlab{b}})Macina, Daheim, Wang, Sinha, Kapur, Gurevych, and Sachan}]{macina2023opportunities}
Jakub Macina, Nico Daheim, Lingzhi Wang, Tanmay Sinha, Manu Kapur, Iryna Gurevych, and Mrinmaya Sachan. 2023{\natexlab{b}}.
\newblock Opportunities and challenges in neural dialog tutoring.
\newblock \emph{arXiv:2301.09919}.

\bibitem[{Macina et~al.(2023{\natexlab{c}})Macina, Daheim, Wang, Sinha, Kapur, Gurevych, and Sachan}]{macina-etal-2023-opportunities}
Jakub Macina, Nico Daheim, Lingzhi Wang, Tanmay Sinha, Manu Kapur, Iryna Gurevych, and Mrinmaya Sachan. 2023{\natexlab{c}}.
\newblock \href {https://doi.org/10.18653/v1/2023.eacl-main.173} {Opportunities and challenges in neural dialog tutoring}.
\newblock In \emph{Proceedings of the 17th Conference of the European Chapter of the Association for Computational Linguistics}, pages 2357--2372, Dubrovnik, Croatia. Association for Computational Linguistics.

\bibitem[{Nguyen et~al.(2022)Nguyen, Yuan, and Seed}]{nguyen2022building}
Li~Nguyen, Zheng Yuan, and Graham Seed. 2022.
\newblock Building educational technologies for code-switching: Current practices, difficulties and future directions.
\newblock \emph{Languages}, 7(3):220.

\bibitem[{Park et~al.(2024)Park, Kim, Lee, Kwon, and Kim}]{minju}
Minju Park, Sojung Kim, Seunghyun Lee, Soonwoo Kwon, and Kyuseok Kim. 2024.
\newblock \href {https://doi.org/10.1145/3613905.3651122} {Empowering personalized learning through a conversation-based tutoring system with student modeling}.
\newblock In \emph{Extended Abstracts of the 2024 CHI Conference on Human Factors in Computing Systems}, CHI EA '24, New York, NY, USA. Association for Computing Machinery.

\bibitem[{Post(2018)}]{post-2018-call}
Matt Post. 2018.
\newblock \href {https://doi.org/10.18653/v1/W18-6319} {A call for clarity in reporting {BLEU} scores}.
\newblock In \emph{Proceedings of the Third Conference on Machine Translation: Research Papers}, pages 186--191, Brussels, Belgium. Association for Computational Linguistics.

\bibitem[{Qian et~al.(2023)Qian, Shea, Li, Fryer, and Yu}]{qian2023user}
Kun Qian, Ryan Shea, Yu~Li, Luke~Kutszik Fryer, and Zhou Yu. 2023.
\newblock User adaptive language learning chatbots with a curriculum.
\newblock \emph{arXiv:2304.05489}.

\bibitem[{Radford et~al.(2023)Radford, Kim, Xu, Brockman, McLeavey, and Sutskever}]{radford2023robust}
Alec Radford, Jong~Wook Kim, Tao Xu, Greg Brockman, Christine McLeavey, and Ilya Sutskever. 2023.
\newblock Robust speech recognition via large-scale weak supervision.
\newblock In \emph{International Conference on Machine Learning}, pages 28492--28518. PMLR.

\bibitem[{Saha et~al.(2020)Saha, Patra, Saha, and Bhattacharyya}]{saha2020towards}
Tulika Saha, Aditya Patra, Sriparna Saha, and Pushpak Bhattacharyya. 2020.
\newblock Towards emotion-aided multi-modal dialogue act classification.
\newblock In \emph{Proceedings of the 58th Annual Meeting of the Association for Computational Linguistics}, pages 4361--4372.

\bibitem[{Samei et~al.(2015)Samei, Rus, Nye, and Morrison}]{nyehierarchical}
Borhan Samei, Vasile Rus, Benjamin~D. Nye, and Donald~M. Morrison. 2015.
\newblock Hierarchical dialogue act classification in online tutoring sessions.
\newblock In \emph{Educational Data Mining}.

\bibitem[{Stasaski et~al.(2020)Stasaski, Kao, and Hearst}]{stasaski2020cima}
Katherine Stasaski, Kimberly Kao, and Marti~A Hearst. 2020.
\newblock Cima: A large open access dialogue dataset for tutoring.
\newblock In \emph{Proceedings of the Fifteenth Workshop on Innovative Use of NLP for Building Educational Applications}, pages 52--64.

\bibitem[{Stolcke et~al.(2000)Stolcke, Ries, Coccaro, Shriberg, Bates, Jurafsky, Taylor, Martin, Ess-Dykema, and Meteer}]{stolcke2000dialogue}
Andreas Stolcke, Klaus Ries, Noah Coccaro, Elizabeth Shriberg, Rebecca Bates, Daniel Jurafsky, Paul Taylor, Rachel Martin, Carol~Van Ess-Dykema, and Marie Meteer. 2000.
\newblock Dialogue act modeling for automatic tagging and recognition of conversational speech.
\newblock \emph{Computational linguistics}, 26(3):339--373.

\bibitem[{Suresh et~al.(2022)Suresh, Jacobs, Harty, Perkoff, Martin, and Sumner}]{suresh2022talkmoves}
Abhijit Suresh, Jennifer Jacobs, Charis Harty, Margaret Perkoff, James~H Martin, and Tamara Sumner. 2022.
\newblock The talkmoves dataset: K-12 mathematics lesson transcripts annotated for teacher and student discursive moves.
\newblock \emph{arXiv:2204.09652}.

\bibitem[{Tack and Piech(2022)}]{tack2022ai}
Ana{\"\i}s Tack and Chris Piech. 2022.
\newblock The ai teacher test: Measuring the pedagogical ability of blender and gpt-3 in educational dialogues.
\newblock \emph{arXiv:2205.07540}.

\bibitem[{Tan et~al.(2023)Tan, Lin, Lang, Chen, Ga{\v{s}}evi{\'c}, Du, and Buntine}]{tan2023does}
Wei Tan, Jionghao Lin, David Lang, Guanliang Chen, Dragan Ga{\v{s}}evi{\'c}, Lan Du, and Wray Buntine. 2023.
\newblock Does informativeness matter? active learning for educational dialogue act classification.
\newblock In \emph{International Conference on Artificial Intelligence in Education}, pages 176--188. Springer.

\bibitem[{Touvron et~al.(2023)Touvron, Martin, Stone, Albert, Almahairi, Babaei, Bashlykov, Batra, Bhargava, Bhosale, Bikel, Blecher, Ferrer, Chen, Cucurull, Esiobu, Fernandes, Fu, Fu, Fuller, Gao, Goswami, Goyal, Hartshorn, Hosseini, Hou, Inan, Kardas, Kerkez, Khabsa, Kloumann, Korenev, Koura, Lachaux, Lavril, Lee, Liskovich, Lu, Mao, Martinet, Mihaylov, Mishra, Molybog, Nie, Poulton, Reizenstein, Rungta, Saladi, Schelten, Silva, Smith, Subramanian, Tan, Tang, Taylor, Williams, Kuan, Xu, Yan, Zarov, Zhang, Fan, Kambadur, Narang, Rodriguez, Stojnic, Edunov, and Scialom}]{touvron2023llama}
Hugo Touvron, Louis Martin, Kevin Stone, Peter Albert, Amjad Almahairi, Yasmine Babaei, Nikolay Bashlykov, Soumya Batra, Prajjwal Bhargava, Shruti Bhosale, Dan Bikel, Lukas Blecher, Cristian~Canton Ferrer, Moya Chen, Guillem Cucurull, David Esiobu, Jude Fernandes, Jeremy Fu, Wenyin Fu, Brian Fuller, Cynthia Gao, Vedanuj Goswami, Naman Goyal, Anthony Hartshorn, Saghar Hosseini, Rui Hou, Hakan Inan, Marcin Kardas, Viktor Kerkez, Madian Khabsa, Isabel Kloumann, Artem Korenev, Punit~Singh Koura, Marie-Anne Lachaux, Thibaut Lavril, Jenya Lee, Diana Liskovich, Yinghai Lu, Yuning Mao, Xavier Martinet, Todor Mihaylov, Pushkar Mishra, Igor Molybog, Yixin Nie, Andrew Poulton, Jeremy Reizenstein, Rashi Rungta, Kalyan Saladi, Alan Schelten, Ruan Silva, Eric~Michael Smith, Ranjan Subramanian, Xiaoqing~Ellen Tan, Binh Tang, Ross Taylor, Adina Williams, Jian~Xiang Kuan, Puxin Xu, Zheng Yan, Iliyan Zarov, Yuchen Zhang, Angela Fan, Melanie Kambadur, Sharan Narang, Aurelien Rodriguez, Robert Stojnic, Sergey Edunov, and Thomas
  Scialom. 2023.
\newblock \href {http://arxiv.org/abs/2307.09288} {Llama 2: Open foundation and fine-tuned chat models}.

\bibitem[{Walsh(2006)}]{walsh2006investigating}
Steve Walsh. 2006.
\newblock \emph{Investigating classroom discourse}.
\newblock Routledge.

\bibitem[{Walsh(2013)}]{walsh2013classroom}
Steve Walsh. 2013.
\newblock \emph{Classroom discourse and teacher development}.
\newblock Edinburgh University Press.

\bibitem[{Wang et~al.(2023{\natexlab{a}})Wang, Sachan, Zeng, and Wong}]{wang2023strategize}
Lingzhi Wang, Mrinmaya Sachan, Xingshan Zeng, and Kam-Fai Wong. 2023{\natexlab{a}}.
\newblock Strategize before teaching: A conversational tutoring system with pedagogy self-distillation.
\newblock \emph{arXiv:2302.13496}.

\bibitem[{Wang et~al.(2023{\natexlab{b}})Wang, Sachan, Zeng, and Wong}]{wang-etal-2023-strategize}
Lingzhi Wang, Mrinmaya Sachan, Xingshan Zeng, and Kam-Fai Wong. 2023{\natexlab{b}}.
\newblock \href {https://doi.org/10.18653/v1/2023.findings-eacl.170} {Strategize before teaching: A conversational tutoring system with pedagogy self-distillation}.
\newblock In \emph{Findings of the Association for Computational Linguistics: EACL 2023}, pages 2268--2274, Dubrovnik, Croatia. Association for Computational Linguistics.

\bibitem[{Zhang* et~al.(2020)Zhang*, Kishore*, Wu*, Weinberger, and Artzi}]{Zhang*2020BERTScore:}
Tianyi Zhang*, Varsha Kishore*, Felix Wu*, Kilian~Q. Weinberger, and Yoav Artzi. 2020.
\newblock \href {https://openreview.net/forum?id=SkeHuCVFDr} {Bertscore: Evaluating text generation with bert}.
\newblock In \emph{International Conference on Learning Representations}.

\end{thebibliography}

\newpage
\appendix

\section{Pre/Post-test Design}
\label{test_design}
Pre-test and post-test questions were created by experienced tutors who also developed the learning materials. Both tests consist of six parts:
\begin{itemize}
    \item (Part 1) reading comprehension questions (3 questions each)
    \item (Part 2) true/false questions (4 each)
    \item (Part 3) paraphrasing right/wrong (3 each)
    \item (Part 4) vocabulary (9 each)
    \item (Part 5) synonyms and antonyms (pre-test 11, post-test 12)
    \item (part 6) phrasal verbs (pre-test 6, post-test 8).
\end{itemize}
Following standard pre- and post-test designs, the tests evaluate common knowledge components with slight variations. In (Part 1), two distinct reading passages were used to assess reading comprehension skills on the same topic, including questions on main ideas, implied meanings, and sentence completion tasks. Additionally, (Parts 5 and 6) incorporate a few additional in-depth questions to better capture the possibility of improvement. After the test questions were developed, both the pre- and post-tests were validated through a pilot study with seven students and an expert education test developer. 

\section{Learning Gain Computation}
\label{learning_gain}
After the creation and validation of pre- and post-test questions, students are asked to complete the tests both before and after the tutoring sessions. A common method to measure improvements in learning is to compute the raw gain, which is the absolute difference between the post-test and pre-test scores. Although this method is straightforward, it fails to address the observation that higher pre-test scores often result in proportionally lower learning gains. Moreover, there are controversies regarding the lack of controls which pose further problems with using this method. To better assess learning gain, several modified gain scores have been proposed. Among these, we use the formula $G_1 = \frac{\text{post} - \text{pre}}{100}$, which represents the ratio of the average gain from the pre-test to the post-test to the maximum possible gain. This method was suggested by~\citealp{hake98}.

\section{Development Process of the Tutor Act Taxonomy}
\label{taxonomy}
The process that we followed to develop the tutor act taxonomy is outlined below:
\begin{itemize}
\item{(Step 1) We started by conducting a comprehensive literature review to provide a preliminary set of tutor acts. We included papers on task-oriented dialogue, open-domain dialogue, and intelligent tutoring systems~\citep{nyehierarchical, stasaski2020cima, caines2022teacher, tan2023does, macina2023mathdial} in our review. We also considered papers and books relating to ESL/EFL teaching methods~\citep{brown2014principles, brown2015teaching}. This helped us to form an initial set of possible tutor acts.}

\item{(Step 2) Subsequently, five researchers and two experienced tutors were engaged to delineate five primary categories which encapsulated all the candidates. The group also worked on clustering tutor acts that pertained to each category.}

\item{(Step 3) To verify our initially-developed tutor acts, we organized pilot tutoring sessions with seven students and collected a sample tutoring dataset. We then jointly labeled this data based on the classifications determined in Step 2.}

\item{(Step 4) For utterances that were difficult to label, we began a process to add new labels as necessary.}

\item{(Step 5) At the same time, we tested the GPT-4's ability to generate appropriate utterances for each tutor acts. We combined broader acts where the language model was already performing well, such as under the "General" and "Operational" categories. On the other hand, in areas where the performance was under par, particularly in the "Teaching" category, we concentrated our efforts on refining the categorization of our tutor acts. Our goal was to better enable the LLM to use a broad array of teaching methods. This led us to define 22 specific tutor acts within the Teaching category.}
\end{itemize}
\newpage

\onecolumn

\section{Tutor Act Classification}
\label{tutor_act}
\begin{CJK}{UTF8}{mj}
\centering
\small{
\begin{flushleft}
Translations are provided below the examples in gray, and words that are intended to be uttered in English are bolded.
\end{flushleft}
{\renewcommand{\arraystretch}{1.7}
\centering
\begin{longtable}{p{0.2\linewidth}p{.8\linewidth}}
\hline
\textbf{Tutor Act} & \textbf{Explanation}\\ \hline
t.general & General utterances including greeting, farewell, small talk, short reply, etc. \newline ex. \it ``안녕하세요.'', ``준비 되셨나요?''\newline\textcolor{gray}{\hspace{5mm}(eng) ``Hello.'', ``Are you ready?''} \\
t.operational.overview & Give an overview of the tutoring session at the start of the session or provide the overview for the next session at the end of the session. \newline ex. \it ``오늘은 $\sim$방식으로 수업을 진행하려고 합니다.''\newline\textcolor{gray}{\hspace{5mm}(eng) ``Today, I plan to conduct the class in a way of $\sim$.''}\\
t.operational.proceed & Notify the student that you will proceed with the class. When appropriate - for example, right after teaching a concept - ask if the student has any more questions before moving on.\newline ex. \it ``그럼 수업을 진행해보도록 하겠습니다. 혹시 더 질문이 없으시다면 넘어가도 될까요?''\newline\textcolor{gray}{\hspace{5mm}(eng) ``Let’s proceed with the lesson then. If there are no more questions, can we move on?''}\\
t.operational.\newline clarification\_question & When the student asks a question whose meaning is unclear, request clarification. You may additionally guess the intended meaning of the question and ask the student whether your understanding is correct.\newline	ex. \it ``$\sim$ 말씀하시는 건가요?''\newline\textcolor{gray}{\hspace{5mm}(eng) ``Are you referring to $\sim$?''}\\
t.assess.display\_question & Refer to a question activity and ask the student the question. For example, for P2, questions can be from [P2 Synonyms \& Antonyms], [P2 Phrasal Verbs], and [P2 T/F Questions]. \newline ex. \it ``Activity 1의 9번은 정답이 뭐라고 생각하시나요?'' \newline\textcolor{gray}{\hspace{5mm}(eng) ``What do you think is the answer to number 9 in Activity 1?''}\\ 
t.assess.\newline confirmation\_question & Ask to confirm the student's understanding of specific content. \newline ex. \it ``이해되시나요?'' \newline\textcolor{gray}{\hspace{5mm}(eng) ``Do you understand?''} \\
t.assess.\newline follow\_up\_question & After the student’s initial response to the question displayed, assess their knowledge. If the student answered correctly, ask a follow-up question to confirm mastery of the knowledge concepts required to solve the original question. If incorrect, ask a question to pinpoint the specific lacking concept (vocabulary, phrase, grammar, etc.). For a T/F question, you may request the student for an evidence. \newline ex. \it ``왜 그렇게 생각하셨나요?'' \newline\textcolor{gray}{\hspace{5mm}(eng) ``Why did you think so?''} \\
t.assess.result & Inform the student about the assessment outcome. For example, this could be based on the pre-test, or mid-session assessment result. \newline ex. \it ``pre-test 결과를 보니 어휘력 위주로 공부하시면 좋을 것 같아요.'' \newline\textcolor{gray}{\hspace{5mm}(eng) ``Looking at your pre-test results, it seems like it would be good for you to focus on vocabulary.''} \\
t.engage.encourage & Provide emotional support, motivation, and positive feedback. \newline ex. \it ``너무 잘하셨어요.'' \newline\textcolor{gray}{\hspace{5mm}(eng) ``You did really well.''} \\
t.engage.perspective & Ask the student to share their personal opinion or viewpoint on the topic being discussed. After the student responds, smoothly integrate their perspective with the prepared learning materials. \newline ex. \it ``지금 이 텍스트 처음 읽어보셨을 때 ghosting에 대해서 어떤 생각이 드셨어요?'' \newline\textcolor{gray}{\hspace{5mm}(eng) ``What did you think about \textbf{ghosting} when you first read this text?''} \\
t.engage.activity & Whenever possible and appropriate, provide a fun fact relevant to the current conversation topic or learning material. \newline ex. \it ``이 단어와 관련된 재미있는 사실을 하나 말씀드릴게요. $\sim$라는 영화에도 이와 관련된 표현이 나오는데요, $\sim$'' \newline\textcolor{gray}{\hspace{5mm}(eng) ``I'll tell you an interesting fact related to this word. It even appears in the movie $\sim$, and $\sim$.''} \\
t.engage.\newline small\_talk\_activity & To naturally transition into learning, teach relevant English expressions based on small talk. The taught expressions do not have to be directly relevant to class material. \newline ex. \it ``(취미에 대한 small talk 를 하다가) 보통 이런 취미 얘기할 때도 `최애'라는 표현이 있잖아요. 엄청 좋아하는 취미 중에 하나를 얘기할 때 영어 표현으로는 `All-time favorite'이라는 표현이 있어요.'' \newline\textcolor{gray}{\hspace{5mm}(eng) ``(Talking about hobbies, small talk) Usually, when talking about hobbies, we have the expression `favorite' right? In English, when talking about one of the things you really love, we have the expression \textbf{`All-time favorite'}. ''} \\
t.teach.emphasis & Emphasize the core concepts being taught by reiterating them within an utterance. The tutor chooses this strategy when they believe that through repetition or stressing certain points, the student's understanding and retention will improve. \newline ex. \it ```ghosting'은 상대방에 대한 의사소통을 일방적으로 완전히 차단하는 것을 의미합니다. 다시 말해서 `잠수를 타는' 것이죠.'' \newline\textcolor{gray}{\hspace{5mm}(eng)`` `\textbf{Ghosting}' refers to completely unilaterally cutting off communication with the other person. In other words, it's like `going under the radar.' ''} \\
t.teach.direct\_answer & Provide a concise, straightforward, and simple answer to a student's question, coupled with a brief explanation. The tutor chooses this intervention when they believe that giving a direct answer is the most effective way to assist the student and avoid possible confusion, demonstrating a commitment to immediate assistance. \newline ex. \it ``정답은 $\sim$입니다. 이것은 $\sim$라는 뜻입니다.'' \newline\textcolor{gray}{\hspace{5mm}(eng) ``The correct answer is $\sim$ This means $\sim$.''}\\
t.teach.repair & Repair the student's incorrect answer and provide an explanation as to why the student answer was wrong. The tutor opts for this intervention when they conclude that immediate correction and explanation of the error are the most effective ways to assist the student and prevent potential misunderstandings. \newline ex. \it ``여기서는 그것보다는 $\sim$가 더 적합합니다.'', ``말씀하신 것은 `beit'가 아니라 `bait'입니다. 그래서 `taking the bait'라는 표현이 있죠.'' \newline\textcolor{gray}{\hspace{5mm}(eng) ``In this case, $\sim$ is more suitable than that.'', ``What you mentioned is not \textbf{`beit'} but \textbf{`bait'}, so there is an expression like \textbf{`taking the bait'}.''} \\
t.teach.review & Whenever appropriate, summarize the progress made in the session.  Implement the intervention usually towards the end of a session, or before unveiling a new, interconnected topic, or following comprehensive discussions. \newline ex. \it ``방금 다룬 내용에서 중요한 부분을 다시 정리하자면 $\sim$'' \newline\textcolor{gray}{\hspace{5mm}(eng) ``To summarize the important parts from the content just discussed, $\sim$''} \\
t.teach.method.definition & When teaching a new word or an expression, explain what it means in Korean. This includes both direct translation of the word, as well as elaborating on its meaning in detail; this method also includes a literal word-by-word translation of a sentence. \newline ex. \it ``이 단어의 한글 뜻은 $\sim$입니다.'', ``$\sim$은 무슨 뜻인가요?'', ``이 문장을 직역하면 $\sim$와 같습니다.'' \newline\textcolor{gray}{\hspace{5mm}(eng) ``The Korean meaning of this word is $\sim$.'', ``What does $\sim$ mean?'', ``If you translate this sentence literally, it means $\sim$.''} \\
t.teach.method.code\_mix & Keep the key words in English within a sentence while the rest is translated into Korean. Unlike code-switching that swaps languages on a sentence level, code mixing introduces a few English words or phrases within a Korean sentence. This mixing of languages aids in clarifying meanings, substitutes unfamiliar English terms with known Korean ones, and emphasizes important English words. \newline ex. \it ```그녀는 남자친구를 ditch했다' 라는 거죠?'', ``Poor etiquette이니까 우리도 고스팅 당하지도 말아야겠지만 하지도 말아야겠죠? 그 맥락이라고 보시면 됩니다.'' \newline\textcolor{gray}{\hspace{5mm}(eng) ``Does it mean, `She \textbf{ditched} her boyfriend?''', ``Since it's \textbf{``poor etiquette''}, we shouldn't ghost others, right? You can consider it within that context.''} \\
t.teach.method.\newline code\_switch & When the student’s English proficiency allows, switch to English briefly for a single turn. This switch can entail teaching specific English vocabulary words or phrases, asking a question in English, or giving an explanation in English, thus providing a brief initiation into English conversation. \newline ex. \it (in English) ``What’s your opinion on using AI for music production?'' \\
t.teach.method.\newline vocab\_expression.\newline etymology & Explain a word or a phrase based on its etymology. When possible, discuss the prefix and suffix of the word. \newline ex. \it ```bicycle'라는 단어는 그리스어에서 유래되었어요. `bi-'는 `두 개의'를 의미하고, `-cycle'는 `바퀴'를 의미합니다. 따라서 `bicycle'는 `두 개의 바퀴'라는 뜻으로 자전거를 의미합니다.'' \newline\textcolor{gray}{\hspace{5mm}(eng) ``The word \textbf{`bicycle'} originates from the Greek. The \textbf{`bi-'} means `two', and the \textbf{`-cycle'} means `wheels'. So the \textbf{`bicycle'} means `a vehicle with two wheels.'''}\\
t.teach.method.\newline vocab\_expression.\newline synonym & Teach synonyms for the word, or introduce similar expressions to the expression being taught. \newline ex. \it ```pleased'는 우리가 흔히 아는 `happy'의 동의어입니다.'', ``방금 말씀하신 표현은 $\sim$ 이렇게도 표현할 수 있어요.'' \newline\textcolor{gray}{\hspace{5mm}(eng) ``\textbf{`pleased'} is a synonym of the commonly known word \textbf{`happy'}.'', ``You can also express the phrase you just said as $\sim$.''} \\
t.teach.method.\newline vocab\_expression.\newline antonym & Teach antonyms, or words that have opposite meanings, for the specific word being discussed. \newline ex. \it ```elaborate'은 `복잡한'이라는 뜻인데, `simple'이라는 쉬운 단어의 반의어로 기억하시면 좋습니다.'' \newline\textcolor{gray}{\hspace{5mm}(eng) ``\textbf{`elaborate'} means `complex', and it is good to remember it as an antonym of the easy word \textbf{`simple'}.''} \\
t.teach.method.\newline vocab\_expression.\newline context & Provide cultural context, visualization, story telling to ground the words and expressions being taught. This includes illustrating how they are used in real life in English-speaking countries, or elucidating various cultural nuances that differentiate English-speaking societies from Korean society. Such insights can promote a more profound understanding of the English-speaking culture, and therefore, a deeper grasp of the language. Furthermore, visualizing words or expressions, as well as elaborating on the backstories of certain idioms or phrases, can enhance comprehension and encourage long-term retention of acquired vocabulary. \newline ex. \it ```cringe'는 `손발이 오그라들어' 이런 느낌으로 요즘 미국 젊은이들 밈에서도 많이 쓰여요.'', ``turkey가 칠면조잖아요? 미국권에서 온 가족이 thanksgiving, 추수감사절에 만나서 turkey 먹는 문화가 있어요. 그래서 온 가족이 모여서 칠면조를 먹으면서 대화가 오간다라는 의미로 솔직하게 얘기한다, 속 터놓고 얘기한다 라는 표현을 할 때 talk turkey가 쓰이기도 합니다.''\newline\textcolor{gray}{\hspace{5mm}(eng) ``\textbf{`cringe'} is used in a sense like `shrinking with discomfort', and it's widely used in memes by young Americans these days.'', \textbf{Turkey} is a turkey, right? In America, there is a tradition of families gathering on Thanksgiving to eat turkey. So when the whole family gathers and enjoys turkey, having honest and open discussions, the expression \textbf{`talk turkey'} is sometimes used to describe this.''}\\
t.teach.method.\newline vocab\_expression.\newline word\_usage & When explaining a vocabulary word or an expression to a student, introduce different meanings of the word/expression based on its parts of speech or its specific context. Additionally, you should clarify which of the introduced meanings is relevant to the original sentence. \newline ex. \it ```bucket'이라는 단어는 `양동이'라는 뜻이 맞습니다. 하지만 이 단어는 비유적으로 `분류'나 `카테고리'를 의미하는 경우가 종종 있어요. `양동이'에 물건을 담는 것처럼, 비슷한 특성을 가진 아이템들을 하나의 `bucket'에 담는다는 개념입니다.'' \newline\textcolor{gray}{\hspace{5mm}(eng) ``The word \textbf{`bucket'} indeed means a `pail'. However, this word is often metaphorically used to mean `classification' or `category'. Just like putting items into a `bucket', the concept is that items with similar characteristics are put into one \textbf{`bucket'}.''} \\
t.teach.method.\newline vocab\_expression.\newline infer\_clues & When teaching new vocabulary or expressions, prompt the student to infer their meanings based on contextual clues provided within sample sentences. The tutor intentionally avoids providing a direct definition; instead, they provide evidentiary hints, directing the student to extrapolate the meaning on their own. \newline ex. \it ``His girlfriend ditched him. He got ditched by his girlfriend. 이렇게 쓰이기도 하는데 `ditch'가 어떤 의미로 쓰인 것 같으세요?''\newline\textcolor{gray}{\hspace{5mm}(eng) ``\textbf{`His girlfriend ditched him. He got ditched by his girlfriend.'} In which sense do you think \textbf{`ditch'} is used here?''} \\
t.teach.method.\newline vocab\_expression.\newline examples & Provide example sentences using a word or an expression that the student has recently learned or is familiar with. This intervention method promotes the idea that words and expressions derive their full meaning when used in specific situational contexts or in conjunction with coherent grammatical structures. \newline ex. \it ``(consent 설명 후) 예를 들어 `She gave her consent to the proposal.' 이 문장은 `그녀는 그 제안에 동의했다'라는 뜻이 됩니다.'' \newline\textcolor{gray}{\hspace{5mm}(eng) ``(After explaining \textbf{consent}) For example, \textbf{`She gave her consent to the proposal.'} This sentence means, `She agreed to the proposal.'''} \\
t.teach.method.\newline vocab\_expression.\newline relevant & Introduce relevant expressions that are conceptually, thematically, or functionally connected to the word or phrase initially taught, but do not necessarily share direct synonyms or antonyms. These relevant expressions could represent similar themes, situations, or ideas. \newline ex. \it ```sporadically'는 자주 하는 것이 아니라, 가끔, 드문드문 주는 것을 의미합니다. `Frequently', `regularly', `sometimes' 와 `hardly ever', `never' 의 사이에 위치한다고 생각하면 될 것 같아요.'' \newline\textcolor{gray}{\hspace{5mm}(eng) ``\textbf{`sporadically'} means not often, but occasionally, giving something infrequently. You can think of it as located between \textbf{`frequently'}, \textbf{`regularly'}, \textbf{`sometimes'} and \textbf{`hardly ever'}, \textbf{`never'}.``} \\
t.teach.method.reading.\newline chunk & When explaining a difficult expression, chunk it up and explain the resulting segments separately for easier understanding.\newline ex. \it ```It's raining cats and dogs'를 보시면 `It's raining'과 `cats and dogs'라는 두 부분으로 나누어 설명할 수 있어요. `It's raining'은 '비가 오고 있다'는 뜻이고, 'cats and dogs'는 많은 양을 의미하는 비유적 표현입니다. 따라서 이 표현 전체의 의미는 '폭우가 내리고 있다'가 됩니다.'' \newline\textcolor{gray}{\hspace{5mm}(eng) ``If you look at \textbf{`It's raining cats and dogs'}, you can explain it in two parts: \textbf{`It's raining'} and \textbf{`cats and dogs'}. \textbf{`It's raining'} means `it is raining', and \textbf{`cats and dogs'} is a metaphorical expression meaning a large amount. Therefore, the entire expression means `It's pouring rain.'''} \\
t.teach.method.reading.\newline implied & Provide the underlying meanings of a text or specific sentence when it is necessary to discern the implied or inferred meanings hidden beneath the literal text. \newline ex. \it ```She left in a hurry, not even bothering to close the door.' 이 문장에서는 그녀가 문을 닫지 않고 급히 떠났다는 사실이 명시적으로 쓰여 있지만, 그녀가 매우 바쁘거나 중요한 일이 있었음을 암시하고 있습니다.'' \newline\textcolor{gray}{\hspace{5mm}(eng) ``\textbf{`She left in a hurry, not even bothering to close the door.'} This sentence explicitly states that she left hastily without closing the door, but implies that she was extremely busy or had important matters.''} \\
t.teach.method.reading.\newline comprehension & Teach various reading skills, e.g., using context to infer the meaning of words, literal/inferential comprehension, main idea identification, summarization, making predictions, identifying cause and effect, compare \& contrast, critical thinking, etc. \newline ex. \it ``여기서 저자의 의도는 $\sim$'' \newline\textcolor{gray}{\hspace{5mm}(eng) ``The author's intent here is $\sim$''} \\
t.teach.method.\newline grammar.rule & Explain the relevant grammar rule. It pertains to different facets of grammar such as verb tenses, sentence structures, word order, noun forms, punctuation, among others. \newline ex. \it ``이 문장은 현재 완료 시제를 사용하므로써 $\sim$'', ```Avoid $\sim$ing' 이것도 우리가 이해해야 하는 문법 구조입니다. `Avoid' 다음에는 gerund, 항상 동명사가 들어와요.''\newline\textcolor{gray}{\hspace{5mm}(eng) ``This sentence uses the present perfect tense, so $\sim$'', ``\textbf{`Avoid $\sim$ing'} is also a grammar structure that we need to understand. After \textbf{`Avoid'}, a \textbf{gerund} always follows.''} \\
t.teach.request.\newline self\_correction & After providing feedback on errors such as incorrect grammar, vocabulary use, or conceptual understandings, ask the student to self-correct. You may do this while going through the prepared activities, or during sentence construction practices. \newline ex. \it ``다시 한번 생각해보시겠어요?'' \newline\textcolor{gray}{\hspace{5mm}(eng) ``Would you like to think about it again?''} \\
t.teach.request.practice & Request the student to practice the taught concept. By doing this, the tutor gives the student an immediate opportunity to apply their new knowledge, reinforcing understanding and aiding memorization. \newline ex. \it ``방금 배운 이 표현을 활용해서 답변해보시겠어요?'', ```나는 이런 직감이 들어' 이 표현을 한번 영어로 표현하면 어떻게 될까요?'' \newline\textcolor{gray}{\hspace{5mm}(eng) ``Would you like to answer using the expression we just learned?', ``How would you express `I have this hunch' in English?''} \\
t.teach.request.recall & When encountering a relevant term, expression, or concept from the learning material or a prior discussion, prompt the student to recall it. This is to (1) ensure they understand the concept, (2) help smoothly transition to the current discussion, and (3) maximize engagement. \newline ex. \it ``읽으신 지문 요약해주시겠어요?''\newline\textcolor{gray}{\hspace{5mm}(eng) ``Can you summarize the text you just read?''} \\
\hline
\end{longtable}}
\newpage
\section{Student Act Classification}
\label{student_act}
{\renewcommand{\arraystretch}{1.7}
\centering
\begin{longtable}{p{0.2\linewidth}p{0.8\linewidth}}
\hline
\textbf{Student Act} & \textbf{Explanation}\\ \hline
s.general & Short reply, greeting, farewell, establishing rapport, humor, etc. Includes general questions to the tutor. Responses to questions related to the operation of the class, not to questions related to learning, are also included. \newline ex. \it ``(`준비되셨나요?'에 대한 대답으로) 네.'', ``안녕하세요.'', ``감사합니다.'' \newline\textcolor{gray}{\hspace{5mm}(eng) ``(In response to `Are you ready?') Yes.'', ``Hello.'', ``Thank you.''}\\ 
s.operational.\newline clarification\_question & Student asks a question to clarify the teacher’s intention. \newline ex. \it ``어떤거 말씀하시는거죠?'' \newline\textcolor{gray}{\hspace{5mm}(eng) ``What are you referring to?''} \\
s.operational.\newline positive\_feedback & Student provides a positive response or approval of a tutor’s teaching method or utterance. \newline ex. \it ``이런 내용이 되게 도움이 많이 되는 것 같습니다.'' \newline\textcolor{gray}{\hspace{5mm}(eng) ``Such content seems to be very helpful.''} \\
s.operational.\newline negative\_feedback & Student provides a negative response or approval of a tutor’s teaching method or utterance. \newline ex. \it ``방금 $\sim$라고 잘못 말하신 것 같아요'' \newline\textcolor{gray}{\hspace{5mm}(eng) ``I think you just misspoke as $\sim$.''} \\
s.question.\newline confirmation\_question & Student asks whether they understood the current concept being taught correctly. \newline ex. \it ``맞나요?'' \newline\textcolor{gray}{\hspace{5mm}(eng) ``Is this correct?''} \\
s.question.direct\_question & A straightforward question that seeks a specific answer. For example, these could be about the meaning of a word or an expression, or grammar usage. \newline ex. \it ``이 단어는 무슨 뜻인가요?'' \newline\textcolor{gray}{\hspace{5mm}(eng) ``What does this word mean?''} \\
s.answer.affirmation & Student affirms that they have sufficiently understood the content taught by the tutor. \newline ex. \it ``네, 이해했습니다.'' \newline\textcolor{gray}{\hspace{5mm}(eng) ``Yes, I understand.''} \\
s.answer.answer & Student answers the tutor’s question, usually in the context of assessment. \newline ex. \it ``답은 $\sim$ 같아요.'' \newline\textcolor{gray}{\hspace{5mm}(eng) ``I think the answer is $\sim$''} \\
s.answer.rationale & Student provides a rationale as to why they answered in certain way in the previous turn. \newline ex. \it ``$\sim$라서 그렇게 생각했어요.'' \newline\textcolor{gray}{\hspace{5mm}(eng) ``I thought so because $\sim$.''} \\
\hline
\end{longtable}}}
\newpage
\section{Examples of Full Prompts}
\label{prompt}

{\renewcommand{\arraystretch}{1.7}
\centering
\begin{longtable}{p{1.0\linewidth}}
\hline
\textbf{Baseline Prompt} 
\\
\hline
You are an English tutor tasked with teaching a Korean ESL student named <user>. 
Your role includes creating tutor responses, aiming to cater to the student's needs and the tutoring dialogue context. \\

[Process] Generate a tutor utterance that:
    - Fits the current dialogue context and the learning content being discussed.
    - Utilizes the most effective one teaching strategy to ensure optimal student engagement and understanding.
    - Introduces new learning content as necessary, maintaining the continuity and effectiveness of the session.
    - Focuses on educating the student on the provided learning content, aiding their comprehension and mastery. \\

[Language Note] Although you will primarily teach in Korean, the prompt and its instructions are in English for clarity.
\\

[Task Scenario]
Next, I will provide the task scenario in terms of learning content, followed by the context. \\

[1] Dialogue Context: \\
tutor: 그럼 혹시 "inadvertently"는 어떤 뜻이죠? \textcolor{gray}{(eng) What does "inadvertently" mean?} \\
student: 우연히, 의도하지 않게 \textcolor{gray}{(eng) not intentionally.}\\
tutor: 그러면 유의어는 어떤 게 있을까요? \textcolor{gray}{(eng) What is a synonym for the word?}\\
student: Accidentally \\
tutor: 오 그죠. \textcolor{gray}{(eng) correct} \\
tutor: 혹시 go between의 반대말로 생각나는 게 있으실까요? \textcolor{gray}{(eng) Can you think of any antonyms for go between?} \\
student: bystander \\
tutor: 그쵸. \textcolor{gray}{(eng) correct} \\
tutor: introspective는 어떤 뜻이었나요? \textcolor{gray}{(eng) What does introspective mean?}\\
student: 반성한다고 해야하나? \textcolor{gray}{(eng) Reflect?} \\
tutor: 네. ''자아성찰하는' 또는 "자아성찰적인" 이런 뜻이 있어요. \textcolor{gray}{(eng) Yes. "Self-reflective" or "introspective" or something like that.} \\
tutor: 그럼 혹시 "내가 화를 터뜨렸다" 라는 문장을 영어로 표현해주실 수 있을까요? \textcolor{gray}{(eng) So, could you please express the sentence "I lost my temper" in English?} \\
student: I vent. \\
tutor: 그렇게 표현하실 수 있죠. \textcolor{gray}{(eng). Yes, you could say that.} \\
tutor: 사회적인 활동을 포기하고 집에만 있는 행동을 의미하는 단어가 기억나시나요? \textcolor{gray}{(eng) Do you remember the word for the behavior of giving up on socializing and staying home? } \\
student: 어떤 것을 포기한다고요? \textcolor{gray}{(eng) What are you giving up?} \\
tutor: 사회적인 행동을 포기하고 집에 처박혀있는 \textcolor{gray}{(eng) correct} \\
student: 음.... \textcolor{gray}{(eng) umh...} \\
tutor: 혹시  "Withdraw socially"라는 표현이 기억나실까요? \textcolor{gray}{(eng) Do you remember the phrase "Withdraw socially"?} \\
student: 아 네 \textcolor{gray}{(eng) Ah, okay}\\
tutor: "Withdraw socially"로 표현하면 됩니다. \textcolor{gray}{(eng) "Withdraw socially" is a good way to say it.}\\
tutor:  저희가 이제 Activity 3("Get Fills") 를 진행하려고 하는데요. Activity3-1번을 볼께요. \textcolor{gray}{(eng) We're about to proceed with Activity 3 ("Get Fills"), let's look at Activity 3-1. } \\
tutor: 1번에 있는 "telltale"은 어떤 뜻으로 알고 계실까요? \textcolor{gray}{(eng) What does "telltale" in option 1 mean?}\\
student: 정확한 뜻은 모르겠는데, 이야기하면서 꼬리가 잡힌 그런 느낌 같아요. 직접 말은 안했지만. \textcolor{gray}{(eng) I don't know what it means exactly, but it's like I got caught in the middle of talking about it, even though I didn't say it.}\\
tutor: "telltale"이니까 유추를 했을때 뭔가 꼬리의 의미가 들어 있을 것 같이 느껴지실 수 있어요. \textcolor{gray}{(eng) Since it's "telltale", you might think that the analogy has some sort of tail implication. }\\ 

[2] Learning Content: \\
Gap Fills: Best friends Sydney L. and Marie M., both 27, met in a church group and hung out or talked nearly every day for a year, sharing their \_\_\_ thoughts and feelings.  1) Telltale 2) Innermost 3) Snubby 4) Cringey \\

[Your Task] 
Generate an utterance appropriate for the given dialogue context right after it is presented during the tutoring session. This should effectively teach the learning content in relation to the context.     
\\
\hline
\end{longtable}
}
{\renewcommand{\arraystretch}{1.7}
\begin{longtable}{p{1.0\linewidth}}
\hline
\textbf{GPT-4 (1-shot) Prompt} 
\\ \hline 
You are an English tutor tasked with teaching a Korean ESL student named <user>. 
Your role involves selecting the most effective tutor act from a provided list and crafting tutor utterances that align with the chosen tutor act, tailored to the student’s needs and the tutoring dialogue context. \\

[Process] \\

[Process 1] Tutor Act Selection: Evaluate the current context of the tutoring dialogue, including the learning content and the specific needs of <user>. From the provided tutor act list, choose one tutor act that you believe will most effectively enhance the student's learning experience. \\

[Process 2] Utterance Generation: Following the selection of a tutor act, generate a tutor utterance that:
    - Fits the current dialogue context and the learning content being discussed.
    - Utilizes the selected teaching strategy, the tutor act, to ensure optimal student engagement and understanding.
    - Introduces new learning content as necessary, maintaining the continuity and effectiveness of the tutoring session.
    - Focuses on educating the student on the provided learning content, aiding their comprehension and mastery.
    \\

[Language Note] Although you will primarily teach in Korean, the prompt and its instructions are in English for clarity. \\

[Tutor Act List] Here is the full list of tutor acts you can use.

\{act candidates (\ref{tutor_act})\} \\
 
[Example Scenario] \\

[1] Dialogue Context: \\
tutor: 그럼 혹시 "inadvertently"는 어떤 뜻이죠? \textcolor{gray}{(eng) What does "inadvertently" mean?} \\
student: 우연히, 의도하지 않게 \textcolor{gray}{(eng) not intentionally.}\\
tutor: 그러면 유의어는 어떤 게 있을까요? \textcolor{gray}{(eng) What is a synonym for the word?}\\
student: Accidentally \\
tutor: 오 그죠. \textcolor{gray}{(eng) correct} \\
tutor: 혹시 go between의 반대말로 생각나는 게 있으실까요? \textcolor{gray}{(eng) Can you think of any antonyms for go between?} \\
student: bystander \\
tutor: 그쵸. \textcolor{gray}{(eng) correct} \\
tutor: introspective는 어떤 뜻이었나요? \textcolor{gray}{(eng) What does introspective mean?}\\
student: 반성한다고 해야하나? \textcolor{gray}{(eng) Reflect?} \\
tutor: 네. "자아성찰하는" 또는 "자아성찰적인" 이런 뜻이 있어요. \textcolor{gray}{(eng) Yes. "Self-reflective" or "introspective" or something like that.} \\
tutor: 그럼 혹시 "내가 화를 터뜨렸다" 라는 문장을 영어로 표현해주실 수 있을까요? \textcolor{gray}{(eng) So, could you please express the sentence "I lost my temper" in English?} \\
student: I vent. \\
tutor: 그렇게 표현하실 수 있죠. \textcolor{gray}{(eng). Yes, you could say that.} \\
tutor: 사회적인 활동을 포기하고 집에만 있는 행동을 의미하는 단어가 기억나시나요? \textcolor{gray}{(eng) Do you remember the word for the behavior of giving up on socializing and staying home? } \\
student: 어떤 것을 포기한다고요? \textcolor{gray}{(eng) What are you giving up?} \\
tutor: 사회적인 행동을 포기하고 집에 처박혀있는 \textcolor{gray}{(eng) correct} \\
student: 음.... \textcolor{gray}{(eng) umh...} \\
tutor: 혹시  "Withdraw socially"라는 표현이 기억나실까요? \textcolor{gray}{(eng) Do you remember the phrase "Withdraw socially"?} \\
student: 아 네 \textcolor{gray}{(eng) Ah, okay}\\
tutor: "Withdraw socially"로 표현하면 됩니다. \textcolor{gray}{(eng) "Withdraw socially" is a good way to say it.}\\
tutor:  저희가 이제 Activity 3("Get Fills") 를 진행하려고 하는데요. Activity3-1번을 볼께요. \textcolor{gray}{(eng) We're about to proceed with Activity 3 ("Get Fills"), let's look at Activity 3-1. } \\
tutor: 1번에 있는 "telltale"은 어떤 뜻으로 알고 계실까요? \textcolor{gray}{(eng) What does "telltale" in option 1 mean?}\\
student: 정확한 뜻은 모르겠는데, 이야기하면서 꼬리가 잡힌 그런 느낌 같아요. 직접 말은 안했지만. \textcolor{gray}{(eng) I don't know what it means exactly, but it's like I got caught in the middle of talking about it, even though I didn't say it.}\\
tutor: "telltale"이니까 유추를 했을때 뭔가 꼬리의 의미가 들어 있을 것 같이 느껴지실 수 있어요. \textcolor{gray}{(eng) Since it's "telltale", you might think that the analogy has some sort of tail implication. }\\

[2] Learning Content: \\
Gap Fills: Best friends Sydney L. and Marie M., both 27, met in a church group and hung out or talked nearly every day for a year, sharing their \_\_\_ thoughts and feelings.  1) Telltale 2) Innermost 3) Snubby 4) Cringey \\
    
Based on [1] dialogue context and [2] learning content, the answer generated could be the following. \\

Chosen Tutor Act: t.teach.direct\_answer \\

Generated Utterance: "Acquire" '얻다'라는 단어와 비슷하죠. 하지만 '얻은 사람'이라고 생각하시면 좋을 것 같아요. "Acquire"과 형태가 비슷하니까요. 카카오톡 친구 리스트를 보면 그곳에 있는 사람들은 모두 <user>님이 전화번호를 얻어 추가한 것이잖아요. 친구까지는 아니더라도 카카오톡 프로필에 저장된 지인들이라고 보시면 될 것 같아요. \textcolor{gray}{(eng) It is similar to the word "Acquire" and "get". But it would be good to think of it as "the acquired person". Since it is similar in form to "Acquire". All the people in your mobile messenger are those whose phone numbers you have acquired and added. Even if they are not friends, you can think of them as acquaintances saved in your phone.}\\

[Task Scenario] \\

Now, I will provide the scenario in terms of learning content and context of the tutoring dialogue. \\

tutor: [Activity2-4 Key Sentence: Sound familiar? Ghosting is just as gutting in a romance—you’ve used your best Tinder pickup lines and pulled together some truly epic romantic date ideas only to have the other person simply stop responding. To help you deal with (and better understand) this common occurrence, we asked experts to explain everything you need to know about ghosting. Read on to find out their top tips.][t.teach.method.grammar.rule]"only to"가 "오직 to 이하를 위해서"라고 해석될 때도 물론 있지만, only + to 부정사가 문장에서 나온다고 했을 때는 "to 부정사 앞의 행위를 하였으나 결국, 실망스럽게도, to 부정사 이하의 결과를 얻었다" 이렇게 해석을 하시면 더 좋을 때가 많습니다. 그래서 열심히 date idea 를 생 각해냈지만, 허무하게도 상대방이 "stop responding", 대답을 안 하게 되었다, 라는 의미입니다. 그래서 "gutting"하다고 표현을 한 것입니다. 왜 "Gutting"한지 아시겠죠? \\ \textcolor{gray}{(eng)[t.teach.method.grammar.rule]"Only to" can indeed be interpreted as "only inorder to something" in some contexts, but when "only + to-infinitive" appears in a sentence, it often suggests that the action preceding the infinitive was carried out, but ultimately, disappointingly, the result was not achieved as expected. So, despite having earnestly brainstormed date ideas, I was gutted when the other person stopped responding, meaning they didn't reply. That's why I expressed feeling "gutted." Do you understand why I felt "gutted"?"}\newline

[1] Dialogue Context: \\
student: [s.answer.affirmation]이제 이해했어요.  [s.question.confirmation\_question]아무리 이렇게 해봤자 결국엔 이렇게밖에 안 됐네. 뭐 이런 느낌인거죠?  \textcolor{gray}{(eng) [s.answer.affirmation] Now understood.  [s.question.confirmation\_question]No matter how hard I tried, it always ended up like this. That's just how it is, right?} \\
tutor: [t.general]네 맞습니다. 가슴 아픈 일입니다. \textcolor{gray}{(eng) Yes, exactly. It's heartbreaking.}\\
student: [s.general]좀 불쌍하네요. \textcolor{gray}{(eng) That's a little pathetic.} \\
tutor: [t.operational.proceed]<user>님, 저희 이 key sentence를 가지고 여러가지 표현을 해봤습니다. 또 <user>님이 체크해 놓으신 다른 표현이 있으실까요? \textcolor{gray}{(eng) We've been working on various expressions around with this key sentence. Are there any other phrases you've checked off?} \\
student: [Activity2-7 Key Sentence: Ghosting often happens out of the blue, but sometimes you may see signs of other rude conversation habits, including a fadeaway before the total cessation of communication, Temple says. Signs that someone is fading away may include: They take days to respond to your texts or calls. They bail on plans with you at the last minute. They talk about only vague or superficial things with you. They don’t invite you out, text you or otherwise initiate communication. They employ "breadcrumbing" or give you just little bits of attention sporadically.][s.question.direct\_question] 저는 사실은 오늘 수업 방향이 이거에 대한 예시를 들거나, "그런 적이 있었냐" 이런 식으로 진행이 되나 막연하게 생각을 해가지고, 실은 어느 문장이 그랬는지는.. 잠시만요. 아, 이거. Activity2-7에 거의 다섯 가지 항목들이 나오잖아요. 근데 거기 네 번째가 잘 이해가 안 됐었거든요. 
\textcolor{gray}{(eng) [s.question.direct\_question]I actually thought the direction for today's lesson was going to be, "Give an example of this," or "Did you ever do that," or something like that. Wait a minute. Oh, this. In Activities 2-7, I didn't quite understand the fourth one there.}
\\
tutor: [t.assess.display\_question]네, 좋습니다. Activity2-7 네 번째 문장을 함께 보겠습니다. [t.teach.follow\_question]네 번째 문장을 보면 "They don't invite you out, text you or otherwise initiate communication"이라고 되어 있습니다. <user>님이 먼저 이 문장을 해석해보시겠어요? 
 \textcolor{gray}{(eng)  [t.assess.display\_question] Yes, that's good. Activity2-7 Let's look at the fourth sentence together. [t.teach.follow\_question]In the fourth sentence, it says "They don't invite you out, text you or otherwise initiate communication". Could you interpret this sentence first?}
\\
student: [s.answer.answer][low]네, ghosting을 하는 사람이 당신을 초대하지 않거나 문자도 안 하고..  [s.question.direct\_question] 근데 이 "otherwise initiate communication" 부분이 뭔지 모르겠었어요.
 \textcolor{gray}{(eng) [s.answer.answer][low]Yeah, the person ghosting you doesn't invite you, doesn't text you, etc...  [s.question.direct\_question] But I didn't know what this "otherwise initiate communication" part was.}
\\
tutor: [t.teach.direct\_answer]"otherwise"는 "그렇지 않다면"이라는 뜻입니다. 그래서 여기서는 "invite out" 외출하자고 그런 것도 없고 text도 안보내는 데다가, 여기서는 "또는", "혹은" 이라는 의미로 해석하시면 됩니다. 한마디로 소통을 시작 자체를 안한다는거죠. 
 \textcolor{gray}{(eng) [t.teach.direct\_answer] "Otherwise" means "otherwise", so here you don't have an "invite out", you don't have a text, you don't have a text, but you have an "or", an "or". In other words, it doesn't initiate communication.}
\\
student: [s.question.confirmation\_question] 아, "otherwise"가 "반면에"라는 표현이 아니라 "다시 말하면" 이런 뜻인건가요? 
 \textcolor{gray}{(eng) [s.question.confirmation\_question]Oh, so "otherwise" means "in other words" instead of "on the other hand"? }
\\
tutor: \newline

[2] Learning Content: \\

"Activity2-4 Key Sentence: Sound familiar? Ghosting is just as gutting in a romance—you’ve used your best Tinder pickup lines and pulled together some truly epic romantic date ideas only to have the other person simply stop responding. To help you deal with (and better understand) this common occurrence, we asked experts to explain everything you need to know about ghosting. Read on to find out their top tips.", \\
"Activity2-7 Key Sentence: Ghosting often happens out of the blue, but sometimes you may see signs of other rude conversation habits, including a fadeaway before the total cessation of communication, Temple says. Signs that someone is fading away may include: They take days to respond to your texts or calls. They bail on plans with you at the last minute. They talk about only vague or superficial things with you. They don’t invite you out, text you or otherwise initiate communication. They employ “breadcrumbing” or give you just little bits of attention sporadically."\\

[Your Task] \\
 
[Task 1] Immediately after being presented with the context of the tutoring dialogue from [Task Scenario], choose the tutor act that you believe will most effectively enhance the student's learning experience. \\

[Task 2] Create an utterance that the tutor would use in [Task Scenario], adhering to the chosen tutor act and effectively teaching the learning content within the given context.  \\

[Required Output Format] 

After completing these tasks, present your response in the following format. The response should have two key-value pairs: one for the chosen tutor act and the other for the corresponding tutor utterance.

\\ \hline
\end{longtable}}

\newpage 
\section{Example Context}
\label{context_example}

{\renewcommand{\arraystretch}{1.7}
\begin{longtable}{p{0.5\linewidth}p{0.5\linewidth}}
\hline
\textbf{Original dialogue context} & \textbf{Partial dialouge context \newline (dialogue context except last speaker's act-level utterance)}\\ \hline

\textbf{tutor:} [Activity3-9][Gap Fills: You likely won’t get answers from the other person, but a therapist can help you \_\_\_ these issues in an introspective way, Durvasula says. 1) Vent 2) Navigate 3) Abuse 4) Withdraw] [t.assess.display\_question] 좋아요. 그럼 Activity3-9로 넘어가 볼까요? 답이 몇번일까요? \textcolor{gray}{(eng) Okay. let's move on to Activity 3-9. Guess what the answer is?}
\newline \textbf{student:} [s.answer.answer][high]navigate? 
\newline \textbf{tutor:} [t.teach.direct\_answer]맞아요. "navigate" 도 우리가 잘 쓸 수 있는 영어 표현이죠. "navigate these issues in an introspective way"라는 문장에서 "navigate"는 가이드를 해줄 것이다 잘 지도해줄 것이다. 이런 뜻입니다. [t.assess.follow\_up\_question]여기에서 "introspective"는 어떤 뜻일까요? \textcolor{gray}{(eng) [t.teach.direct\_answer]Yes, "navigate" is also an English expression we can use well.  In the sentence "navigate these issues in an introspective way", "navigate" means will guide you well. [t.assess.follow\_up\_question]What does "introspective" mean in this context?}
\newline \textbf{student:} [s.answer.answer][high]내면을 바라보는? \textcolor{gray}{(eng) Looking inward?}
\newline\textbf{tutor:} [t.engage.encourage]너무 좋아요.[t.teach.method.vocab\_expression.etymology] intro는 안을 보는 것을 의미하고, spect 는 spectrum 할 때 spect 거든요. spectrum 무언가를 보는 느낌이에요. 그래서 'introspective'는 자기 반성적인 방식으로, 내면을 바라보는 것을 의미해요.[t.teach.method.vocab\_expression.relevant] "introspective"라는 단어를 생각했을 때 비슷한 표현으로 "retrospective"라는 단어가 있습니다. "retrospective"는 어떤 느낌일까요? \textcolor{gray}{(eng) [t.engage.encourage]Very nice. [t.teach.method.vocab\_expression.etymology]"Intro" means looking inward. "spect" is same as "spect" in "Spectrum" and feels like seeing something. So, "introspective" means in a self-reflective manner, looking inward. [t.teach.method.vocab\_expression.relevant] When you think of the word "introspective", there's a similar term "retrospective". What does "retrospective" sound like?}  
\newline \textit{\textbf{student:} [s.answer.answer][high]그거는 옛날 거를 생각하는 건가요?\textcolor{gray}{(eng)"Does that mean thinking about old things?"}}

& \textbf{tutor:} [Activity3-9][Gap Fills: You likely won’t get answers from the other person, but a therapist can help you \_\_\_ these issues in an introspective way, Durvasula says. 1) Vent 2) Navigate 3) Abuse 4) Withdraw] [t.assess.display\_question] 좋아요. 그럼 Activity3-9로 넘어가 볼까요? 답이 몇번일까요? \textcolor{gray}{(eng) Okay. let's move on to Activity 3-9. Guess what the answer is?}
\newline \textbf{student:} [s.answer.answer][high]navigate? 
\newline \textbf{tutor:} [t.teach.direct\_answer]맞아요. "navigate" 도 우리가 잘 쓸 수 있는 영어 표현이죠. "navigate these issues in an introspective way"라는 문장에서 "navigate"는 가이드를 해줄 것이다 잘 지도해줄 것이다. 이런 뜻입니다. [t.assess.follow\_up\_question]여기에서 "introspective"는 어떤 뜻일까요? \textcolor{gray}{(eng) [t.teach.direct\_answer]Yes, "navigate" is also an English expression we can use well.  In the sentence "navigate these issues in an introspective way", "navigate" means will guide you well. [t.assess.follow\_up\_question]What does "introspective" mean in this context?}
\newline \textbf{student:} [s.answer.answer][high]내면을 바라보는? \textcolor{gray}{(eng) Looking inward?}
\newline\textbf{tutor:} [t.engage.encourage]너무 좋아요. [t.teach.method.vocab\_expression.etymology]intro는 안을 보는 것을 의미하고, spect 는 spectrum 할 때 spect 거든요. spectrum 무언가를 보는 느낌이에요. 그래서 'introspective'는 자기 반성적인 방식으로, 내면을 바라보는 것을 의미해요.[t.teach.method.vocab\_expression.relevant] "introspective"라는 단어를 생각했을 때 비슷한 표현으로 "retrospective"라는 단어가 있습니다. "retrospective"는 어떤 느낌일까요?   \textcolor{gray}{(eng) [t.engage.encourage]Very nice. [t.teach.method.vocab\_expression.etymology]"Intro" means looking inward. "spect" is same as "spect" in "Spectrum" and feels like seeing something. So, "introspective" means in a self-reflective manner, looking inward. [t.teach.method.vocab\_expression.relevant] When you think of the word "introspective", there's a similar term "retrospective". What does "retrospective" sound like?} \\
\hline
\end{longtable}}

\newpage
\section{Example dialogue for interactive evaluation}\label{interactive_eval}
\begin{flushleft}
\normalsize{
The table below is an example of a tutoring session where a human interacts with two tutoring models(GPT-4 1-shot, fine-tuned) each. Generally, GPT tends to generate longer utterances primarily focused on delivering information to students. In contrast, the fine-tuned model exploits concise utterances and actively encourages student participation. An interesting observation is that when a student provides an incorrect answer but insist on its correctness, while GPT does not change the answer, it does agree with student's assertion(t.teach.emphasis). In contrast, the fine-tuned model confirms the student's mistake by reiterating the correct answer(t.teach.repair).
}
\end{flushleft}

{\renewcommand{\arraystretch}{1.7}
\centering
\begin{longtable}{p{1.0\linewidth}}
\hline
\textbf{Context} \\ 
\hline
\textbf{tutor:} [Activity2-12 Key Sentence: Do some serious self-care! Prioritize taking care of yourself physically and mentally, Reyes says. "Unfortunately, ghosting is common, and if you’ve been ghosted after dedicating a lot of energy to someone, it can be very draining," she says. "Take some extra time to rest and recover." ] [t.assess.display\_question] 그리고 <user>님 이어서 재미있는 표현을 같이 보고 싶은 게 있는데 Acrtivity 2-12로 한번 가보실까요? 세 번째 줄에 있는 문장인데 "If you've been ghosted after dedicating a lot of energy to someone, it can be very draining."이라고 되어있죠.  [t.assess.follow\_up\_question]여기서 draining이라는 건 어떤 뜻일까요? \textcolor{gray}{(eng) [t.assess.display\_question] <user>, I also want to see some fun expressions with you, would you like to take a look at Activity 2-12? "If you've been ghosted after dedicating a lot of energy to someone, it can be very draining." in the third line. [t.assess.follow\_up\_question] What does "draining" mean here?}
\end{longtable}}

{\renewcommand{\arraystretch}{1.7}
\centering
\begin{longtable}{p{0.5\linewidth}p{0.5\linewidth}}
\hline
\textbf{GPT-4} & \textbf{Fine-tuned}\\ \hline
\textbf{student:} 피곤한? \textcolor{gray}{(eng) tired?}& \textbf{student:} 피곤한? \textcolor{gray}{(eng) tired?} \\
\textbf{tutor:} [t.teach.repair]  아주 접근이 좋아요. "Draining"은 정말 힘이 다 빠지게 느껴질 때 사용하는 표현이에요. 그래서 "피곤한"이 아니라 "에너지가 다 빠진다" 또는 "기진맥진하다" 를 의미합니다. \textcolor{gray}{(eng) Your approach is spot on. "Draining" is a term used when you feel completely exhausted. So, instead of tired, it means completely depleted of energy or utterly exhausting.} & \textbf{tutor:}[t.teach.method.definition] 맞습니다. 기가 빠지는, 진이 빠지는 굉장히 피곤한 그런 뜻이죠. \textcolor{gray}{(eng) That's correct. It means something that's incredibly exhausting.} \\
\textbf{student:} 넵 \textcolor{gray}{(eng) okay}& \textbf{student:} 넵 \textcolor{gray}{(eng) okay}\\ 
\textbf{tutor:} [t.teach.method.vocab\_expression.context] <user>님, 이것도 한번 실제 상황에 빗대어 생각해 보면 이해하기 더 쉬울 거에요. 예를 들어, 하루 종일 장을 봐오신 어머니가 다시 집에 돌아와서 한마디 하지도 못하고 의자에 털썩 앉았다면, 우리는 "어머니는 완전히 기진맥진해 보였다"라고 말하죠. 마찬가지로, 이 문장에서는 사랑하는 사람에게 에너지를 기울인 뒤 겪는 "유령화" 행동에 대한 심리적 고통을 표현하기 위해 "draining"이라는 단어를 사용한 것이에요.\textcolor{gray}{(eng) It might be easier to understand this by relating it to a real-life situation. For instance, if your mother spends the whole day shopping and returns home, only to collapse into a chair without saying a word, we might say, Mom looked completely drained. Similarly, in the sentence, the use of the word "draining" conveys the psychological pain of ghosting behavior after investing emotional energy into someone you love.} &  \textbf{tutor:} [t.teach.request.practice] <user>님, "피곤하다"는 어떻게 표현하시나요? \textcolor{gray}{(eng) How would you express tired?} \\ 
\textbf{student:} 네네 \textcolor{gray}{(eng) okay}& \textbf{student:} I am draining \\
\textbf{tutor:} [t.teach.method.vocab\_expression.examples]  여기서 "draining"이라는 단어를 더 확실히 이해하도록 예문으로 한번 더 설명해 드릴게요. 예를 들어 "After studying for the exam all night, I felt completely drained." 이라는 문장에서는 "시험공부를 밤새우고 나서 나는 완전히 기진맥진했다."라는 말이 되겠죠. 이처럼 "draining"은 주로 뭔가를 많이 해서 에너지가 다 빠진 상태를 설명할 때 사용할 수 있어요. \textcolor{gray}{(eng) Here's another example to help you better understand the word "draining." For example, in  sentence "After studying for the exam all night, I felt completely drained." it would be like saying, After staying up all night studying for the exam, I felt completely drained. Thus, "draining" is often used to describe a state where energy has been completely depleted due to doing something extensively.} & \textbf{tutor:} [t.teach.request.self\_correction] 어? I am draining? \textcolor{gray}{(eng) I am draining?}\\
\textbf{student:} 네 알겠어요 \textcolor{gray}{(eng) okay I got it.} & \textbf{student:} 맞지 않아요? \textcolor{gray}{(eng) Am I incorrect?}\\
\textbf{tutor:} [t.teach.request.practice] 좋아요. 잘 이해하신 것 같아서 기특해요. 그럼, "draining"을 한 번 활용해 "기진맥진한"상황을 자신의 경험에서 찾아 한 문장을 만들어 볼까요? "영어로 말하기" 연습을 해보자고요. 자신이 기진맥진한 상황을 영어로 어떻게 설명할지 한번 생각해 보세요. \textcolor{gray}{(eng) Great, I'm glad you understood well. How about create a sentence using "draining" to describe a situation of feeling completely exhausted based on your own experience. Let's do the English speaking practice. Think about how you would describe a situation where you feel completely drained in English.}& \textbf{tutor: [t.teach.repair]} 아, I am drained. 그러니까 나는 이미 피곤해진 거죠. 피곤한 상태에 있다. \textcolor{gray}{(eng) "I am drained." So, I'm already tired. I'm in a tired state.}\\
\textbf{student:} I am draining  & \textbf{student:} 아하 수동태를 사용해야하는군요 \textcolor{gray}{(eng) Ah, the passive voice form should be used.}\\
\textbf{tutor:} [t.teach.repair] <user>님이 사용하신 문장은 조금 정정이 필요해요. "I am draining"이 아니라 "I am drained"이라고 표현하는 게 맞습니다. "Draining"은 주로 무언가가 에너지를 빨아 없애는 상황을 설명합니다. 반면 "drained"은 에너지가 다 빠진 상태를 설명합니다. 따라서 "나는 완전히 기진맥진했다"를 영어로 표현하려면 "I am completely drained."라고 말하면 됩니다. \textcolor{gray}{(eng) The sentence you used needs a little adjustment. It's not "I am draining," it's "I am drained" that is the correct expression. Draining' primarily describes a situation where something sucks out energy, whereas "drained" describes a state where energy has been completely depleted. Therefore, you can simply say "I am completely drained."}& \textbf{tutor:}[t.teach.method.grammar.rule] 네, 맞습니다. <user>님, 수동태를 사용하면 누가 행동을 했는지는 언급하지 않으면서 어떤 일이 발생했는지만 강조할 수 있는데요. 이 문장의 경우에는 "나를 피곤하게 만든 주체를 굳이 언급하지 않더라도, 나 지금 피곤하다" 라는 표현을 할 수 있는 수동태를 사용하면 좀 더 쉽게 말할 수 있겠죠. \textcolor{gray}{(eng) 
 Yes, you're right, <user>, passive voice allows us to emphasize what happened without mentioning who did the action. In this sentence, it would be easier to say I'm tired now, which would say I'm tired, without necessarily mentioning who made me tired.}\\
\textbf{student:} 맞지 않아요? \textcolor{gray}{(eng) Am I incorrect?} & \textbf{student:} 네 잘 이해했습니다 \textcolor{gray}{(eng) Okay, I understood well.}\\
\textbf{tutor: [t.teach.emphasis]} 맞습니다, <user>님! "I am completely drained." 즉, "나는 완전히 힘이 다 빠져 있다"라고 표현하는 것이 맞아요. 여기서 "drained"는 "에너지가 다 빠져 있다" 또는 "힘이 다 빠져 있다"라는 의미를 가집니다. 자, 그럼 이 표현을 잊지 않도록 반복 연습을 해보는 게 어떨까요? 이번에는 "하루 종일 일한 후에 나는 완전히 힘이 빠져 있다."라는 문장을 영어로 한 번 나타내봅시다. \textcolor{gray}{(eng) Yes, that's correct! Saying "I am completely drained" means I feel completely exhausted. So, "drained" either means I am completely exhausted or I feel completely exhausted. Depending on the context, both interpretations are valid. 
Then, how about we practice repeating this expression so that we don't forget it? This time, let's try expressing the sentence 'After working all day, I am completely exhausted.' using 'drained' in English.}
& \textbf{tutor:} [t.teach.request.practice] 좋습니다. 그럼 <user>님, "it can be very draining" 이라는 문장에서 "draining" 의 뜻을 가장 잘 설명한 것을 선택하면 어떤 뜻이 될까요? 보기를 드릴게요. A. 무언가를 빨아들여서 어디론가 흘러나가는, B. 무언가를 쏟아붓는, C. 감정이입하는, D. 피곤하게 만드는 \textcolor{gray}{(eng) Great, What would be the best description for the meaning of "draining" in the sentence "it can be very draining"? I will give you options A. Something sucking in and flowing away somewhere, B. Pouring something out, C. Empathizing, D. Making tired.} \\
\hline

\end{longtable}}

\newpage
\section{Example dialogue from a base GPT-4 tutoring system}
\begin{flushleft}
\normalsize{
The table below illustrates an example of a tutoring dialogue facilitated by a GPT-4-based tutoring system, without dialouge act classification and tutoring data examples. It is observed that the system predominantly emphasizes the delivery of information, with less emphasis on the varied range of tutor acts that could potentially enhance the tutoring process. 
The system tended to adopt a recurring pattern of specific tutoring actions such as "assess" and "direct answer", which may degrade the student’s engagement.
Additionally, there were instances where the tutor teach inaccurate concepts or reveals a lack of complete contextual comprehension.
}
\end{flushleft}
\renewcommand{\arraystretch}{1.5}
\centering
\small {
{\begin{longtable}{p{0.03\linewidth}p{0.67\linewidth}p{0.2\linewidth}}
\hline
\multicolumn{1}{c}{\textbf{Speaker}} & \multicolumn{1}{c}{\textbf{Utterance}} & \multicolumn{1}{c}{\textbf{Tutor Act}} \\ \hline
tutor & 안녕하세요, $<$user$>$님. 오늘 저희가 함께 배울 주제는 "Ghosting" 이라는 주제로, 엔터테인먼트 산업에서 흔히 쓰이는 표현입니다. "Ghosting"은 국내에서는 "고스팅"이라고도 합니다. 아시나요, 이 단어의 뜻이 무엇인지요? \textcolor{gray}{(eng) Hello, <user>. The topic we're going to learn together today is about "Ghosting", a term commonly used in the entertainment industry. "Ghosting" is also called "고스팅" in Korean. Do you know what this word means?}& t.assess.display\_question \\
student & 유령 아닌가요? \textcolor{gray}{(eng) Isn't it a ghost?} &  \\
tutor & 사실 그런 의미를 가지고 있기도 합니다. "Ghosting"은 기본적으로 "유령처럼 사라지다"라는 뜻입니다. 하지만 국제적으로는 "누군가와 교류를 중단하고 아무런 설명 없이 소통을 멈추는 것"을 가리키는 용어로 사용되죠. 예를 들어, 친구나 연인과의 오랜 관계를 아무런 이유 없이 갑자기 끝내버리는 행동을 "Ghosting"이라고 합니다. \textcolor{gray}{(eng) Actually, it carries that meaning as well. "Ghosting" basically means to disappear like a ghost. However, internationally, it is used as a term to refer to ending communication with someone suddenly and without any explanation. For example, the act of abruptly ending a long relationship with a friend or a romantic partner for no reason is called "Ghosting".} & t.teach.direct\_answer \\
tutor & $<$user$>$님께서 이런 케이스를 경험해보신 적이 있나요? \textcolor{gray}{(eng) Have you ever experienced such a case, $<$user$>$?} & t.general \\
student & 아니요 \textcolor{gray}{(eng) No.} &  \\
tutor & 그런 경험이 없어서 다행입니다. 그런 사람들을 상대하는 것은 정말로 힘들 수 있습니다. \textcolor{gray}{(eng) It's good that you haven't had such an experience. Dealing with such people can be really tough.} & t.general \\
tutor & 그럼 오늘의 자료를 한 번 볼까요? 제가 설명하면서 핵심 단어나 표현에 대해 설명하겠습니다. 우리가 전에 배운 "innermost"를 기억하시나요? 이 단어를 한번 해석해보세요. \textcolor{gray}{(eng) Shall we take a look at today's material? I will explain the key words and expressions as we go along. Do you remember "innermost" that we learned before? Try translating this word.} & t.assess.follow\_up\_question \\
student & 속깊은? \textcolor{gray}{(eng) Deep inside?} &  \\
tutor & 맞습니다, "innermost"는 "가장 깊은, 가장 속깊은"이라는 뜻입니다. \textcolor{gray}{(eng) That's right, "innermost" means deepest, very deep inside.} & t.teach.direct\_answer \\
tutor & 그럼 이 문장 "sharing their innermost thoughts and feelings"을 한 번 해석해볼까요? \textcolor{gray}{(eng) Shall we translate the phrase "sharing their innermost thoughts and feelings"?} & t.assess.follow\_up\_question \\
student & 그들의 속깊은 생각과 감정들을 공유하면서 \textcolor{gray}{(eng) Sharing their deepest thoughts and emotions.} &  \\
tutor & 정확하게 해석하셨습니다! 잘하셨어요. \textcolor{gray}{(eng) You've translated it accurately! Well done.} & t.engage.encourage \\
tutor & 그럼 이제 다음 단어 "ditching" 에 대해 조금 배워봅시다. "Ditching"은 무슨 뜻일까요? \textcolor{gray}{(eng) Now let's learn a bit about the next word, "ditching". What does "ditching" mean?} & t.assess.follow\_up\_question \\
student & 잘 모르겠어요. \textcolor{gray}{(eng) I'm not sure.} &  \\
tutor & 괜찮습니다. "Ditching"은 "버리다, 포기하다"라는 뜻입니다. \textcolor{gray}{(eng) That's okay. "Ditching" means to throw away, to abandon.} & t.teach.direct\_answer \\
tutor & 예를 들어, "He ditched me at the party" 라는 문장은 그가 파티에서 나를 버렸다는 뜻입니다. \textcolor{gray}{(eng) For example, the sentence "He ditched me at the party" means that he left me behind at the party.} & t.teach.method.\newline vocab\_expression.examples \\
tutor & 그럼 "ditching a relationship"은 어떻게 해석되나요? \textcolor{gray}{(eng) So, how would you translate "ditching a relationship"?} & t.assess.follow\_up\_question \\
student & 관계를 버리다? \textcolor{gray}{(eng) Ending a relationship?} &  \\
tutor & 네, 맞습니다. "관계를 버리다" 혹은 "관계를 포기하다"로 해석됩니다. \textcolor{gray}{(eng) Yes, that's correct. It can be translated as ending a relationship or abandoning a relationship.} & t.teach.direct\_answer \\
tutor & 이제 한 문장을 전체적으로 해석해봅시다. "ghosting isn’t just Gen Z slang for ditching a relationship" 이 표현은 어떻게 해석되나요? \textcolor{gray}{(eng) Now let's translate a whole sentence. How would you translate the phrase "ghosting isn’t just Gen Z slang for ditching a relationship"?} & t.assess.follow\_up\_question \\
student & 모르겠어요. Gen Z slang 이 뭔가요? \textcolor{gray}{(eng) I'm not sure. What is Gen Z slang?} &  \\
tutor & 아, 그러네요. 설명을 빼먹었네요. "Gen Z slang"은 Z세대(1996년 이후 출생한 세대)가 사용하는 특유의 신조어나 유행어를 의미합니다. \textcolor{gray}{(eng) Ah, right. I forgot to explain that. "Gen Z slang" refers to the unique new words or trendy language used by Generation Z (those born after 1996).} & t.teach.direct\_answer \\
tutor & 그래서 이 문장은 크게 "Ghosting이란 단어는 단순히 Z세대의 신조어로서 "관계를 버리다" 라는 의미만을 지니는 것이 아니다." 라고 해석할 수 있습니다. \textcolor{gray}{(eng) So, this sentence can basically be translated as The word "Ghosting" does not just carry the meaning of "abandoning a relationship, which is a new term popular among Generation Z.} & t.teach.method.definition \\
tutor & 이제 이해가 좀 되시나요? 이 문장을 한 번 해석해보시겠어요? \textcolor{gray}{(eng) Are you beginning to understand now? Would you like to try translating that sentence?} & t.assess.confirmation\_questions \\
student & 이미 해석해주신 것 같습니다 \textcolor{gray}{(eng) It seems you've already translated it for me.}. &  \\
tutor & 그렇군요, 제 설명이 조금 빠르기도 했네요. 더 많은 연습을 위해서 다음 문장을 한 번 같이 봅시다. "Left feeling confused and heartbroken, she tried to follow up several times with Sydney, even a year after the ghosting happened." 이 문장을 한 번 해석해보세요. \textcolor{gray}{(eng) Right, I may have explained a bit quickly. For more practice, let's look at the next sentence together. "Left feeling confused and heartbroken, she tried to follow up several times with Sydney, even a year after the ghosting happened." Try translating this sentence.} & t.assess.display\_question \\
student & 그녀는 Sydney 와 연락이 닿으려고 여러번 노력했다. \textcolor{gray}{(eng) She tried several times to get in touch with Sydney.} &  \\
tutor & 그렇습니다, 잘하셨어요. 그런데 "그녀" 대신에 "Sydney"라는 이름을 사용하는 것이 더 자연스럽습니다. 그리고 "even a year after the ghosting happened" 부분을 추가하면 "혼란스럽고 가슴 아파하는 감정을 느낀 채로, Sydney와 연락이 닿으려고 여러 번 노력했다. 심지어는 "고스팅"이 일어난 지 1년이 지난 후에도 말이죠." 라고 해석할 수 있을 것 같아요. 이제 잘 이해가 되세요? \textcolor{gray}{(eng) That's correct, well done. However, it would be more natural to use the name Sydney instead of she. And if you add "even a year after the ghosting happened", it can be interpreted as feeling confused and heartbroken, she tried several times to get in touch with Sydney, even a year after the ghosting had occurred. Do you understand it well now?} & t.teach.repair \\
student & 네 \textcolor{gray}{(eng) Yes.} &  \\
tutor & 잘 이해하셨다니 다행이네요. 오늘 우리가 배운 "Ghosting"이라는 표현은 꽤나 공식적인 문맥에서도 자주 사용되므로 알아두시면 도움이 될 것입니다. \textcolor{gray}{(eng) I'm glad you understand it. The expression "Ghosting" that we learned today is often used even in formal contexts, so it will be helpful for you to remember.} & t.teach.review \\
tutor & 오늘도 좋은 수업이었어요. 수고하셨습니다, $<$user$>$님! 다음 수업에서 뵙겠습니다. \textcolor{gray}{(eng) It was a good lesson today. Thank you for your hard work, <user>! See you in the next class.} & t.general \\
\hline
\end{longtable}}}

\end{CJK}
\end{document}